\documentclass[journal]{IEEEtran}
\IEEEoverridecommandlockouts
\usepackage{graphicx}
\def\BibTeX{{\rm B\kern-.05em{\sc i\kern-.025em b}\kern-.08em
    T\kern-.1667em\lower.7ex\hbox{E}\kern-.125emX}}

\usepackage{blindtext}
\usepackage{caption}

\let\oldtwocolumn\twocolumn
\renewcommand\twocolumn[1][]{%
    \oldtwocolumn[{#1}{
    \begin{center}
    \vskip-5ex
        \centering
        \includegraphics[width=0.99\textwidth]{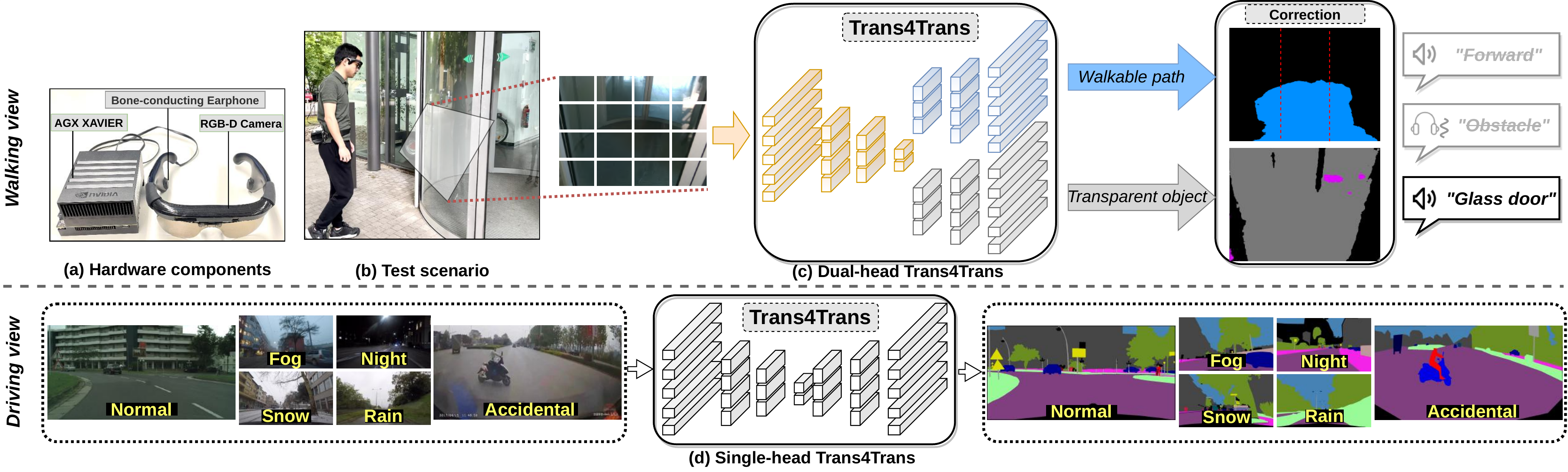}
        \captionof{figure}{\small (a) the assistive system 
        equipped with smart vision glasses and a portable GPU processor
        is tested (b) in front of a glass door. The input image is segmented as \textbf{\textcolor{blue}{\emph{walkable path}}} and \textbf{\textcolor{gray}{\emph{glass door}}} by (c) our Transformer for Transparency~(Trans4Trans) model, which are safety-critical for navigation. The user interface consists of vibration and voice feedback. After training on normal and adverse data, our (d) Trans4Trans model reaches high robustness when applied to various real-world driving scenarios.}
        \label{fig1:sys}
    \end{center}
    }]
}

\usepackage{amsmath,amssymb,amsfonts,bm}
\usepackage{lipsum}
\usepackage{times}
\usepackage{epsfig}

\usepackage{stfloats}
\usepackage{multicol}
\usepackage[inline]{enumitem}
\usepackage{algorithmic}
\usepackage{graphicx}
\usepackage{textcomp}

\usepackage{booktabs}
\usepackage{multirow}
\usepackage{hyperref}
\hypersetup{
colorlinks=true,linkcolor=red,citecolor=rblue,urlcolor=black}

\usepackage[utf8]{inputenc}
\usepackage[T1]{fontenc}
\usepackage{float}
\usepackage{adjustbox}
\usepackage{array}
\usepackage{etoolbox}
\usepackage{dsfont}

\usepackage{paralist}
\usepackage[ruled,linesnumbered]{algorithm2e}

\usepackage{tabulary}
\usepackage[table]{xcolor}

\definecolor{gray}{rgb}{0.3,0.3,0.3}
\definecolor{blue}{rgb}{0,0.5,1}
\definecolor{mask_red}{rgb}{1,0,0.8}
\definecolor{green}{rgb}{0.2,1,0.2}
\definecolor{rblue}{rgb}{0,0,1}
\newcommand{\gray}[1]{\textcolor{gray}{#1}}
\newcommand{\green}[1]{\textcolor[RGB]{96,177,87}{#1}}
\newcommand{\fn}[1]{\footnotesize{#1}}
\newcommand{\gbf}[1]{\green{\bf{\fn{(#1)}}}}
\newcommand{\rbf}[1]{\gray{\bf{\fn{(#1)}}}}

\makeatletter
\DeclareRobustCommand\onedot{\futurelet\@let@token\@onedot}
\def\@onedot{\ifx\@let@token.\else.\null\fi\xspace}

\def\eg{\emph{e.g}\onedot} 
\def\ie{\emph{i.e}\onedot}

\def\wrt{w.r.t\onedot} 
\def\etal{\emph{et al}\onedot}

\usepackage{subcaption}

\usepackage[capitalise]{cleveref} 

\begin{document}

\title{Trans4Trans: Efficient Transformer for Transparent Object and Semantic Scene Segmentation in Real-World Navigation Assistance}
\author{Jiaming Zhang$^{1}$, Kailun Yang$^{1}$, Angela Constantinescu$^{2}$, Kunyu Peng$^{1}$, Karin Müller$^{2}$, and Rainer Stiefelhagen$^{1,2}$
\thanks{This work was supported in part through the AccessibleMaps project by the Federal Ministry of Labor and Social Affairs (BMAS) under the Grant No. 01KM151112, in part by the University of Excellence through the ``KIT Future Fields'' project, and in part by Hangzhou SurImage Company Ltd.
\emph{(Corresponding author: Kailun Yang.)}}
\thanks{$^{1}$Authors are with Computer Vision for Human-Computer Interaction Lab, and $^{2}$authors are with Center for Digital Accessibility and Assistive Technology, Karlsruhe Institute of Technology, Germany (e-mail: jiaming.zhang@kit.edu, kailun.yang@kit.edu, angela.constantinescu@kit.edu, kunyu.peng@kit.edu, karin.mueller2@kit.edu, rainer.stiefelhagen@kit.edu).}
\thanks{Code will be made publicly available at: \url{https://github.com/jamycheung/Trans4Trans}.}
}

\maketitle
\bstctlcite{IEEEexample:BSTcontrol}
\begin{abstract}
Transparent objects, such as glass walls and doors, constitute architectural obstacles hindering the mobility of people with low vision or blindness. For instance, the open space behind glass doors is inaccessible,
unless it is correctly perceived and interacted with. However, traditional assistive technologies rarely cover the segmentation of these safety-critical transparent objects. In this paper, we build a wearable system with a novel dual-head Transformer for Transparency \emph{(Trans4Trans)} perception model, which can segment general- and transparent objects. The two dense segmentation results are further combined with depth information in the system to help users navigate safely and assist them to negotiate transparent obstacles. We propose a lightweight Transformer Parsing Module \emph{(TPM)} to perform multi-scale feature interpretation in the transformer-based decoder. Benefiting from TPM, the double decoders can perform joint learning from corresponding datasets to pursue robustness, meanwhile maintain efficiency on a portable GPU, with negligible calculation increase. The entire Trans4Trans model is constructed in a symmetrical encoder-decoder architecture, which outperforms state-of-the-art methods on the test sets of Stanford2D3D and Trans10K-v2 datasets, obtaining mIoU of $45.13\%$ and $75.14\%$, respectively. Through a user study and various pre-tests conducted in indoor and outdoor scenes, the usability and reliability of our assistive system have been extensively verified. Meanwhile, the Tran4Trans model has outstanding performances on driving scene datasets. On Cityscapes, ACDC, and DADA-seg datasets corresponding to common environments, adverse weather, and traffic accident scenarios, mIoU scores of $81.5\%$, $76.3\%$, and $39.2\%$ are obtained, demonstrating its high efficiency and robustness for real-world transportation applications.

\end{abstract}

\begin{IEEEkeywords} 
Computer vision for the visually impaired, wearable assistive system, transparent object segmentation, semantic segmentation, scene understanding
\end{IEEEkeywords}

\IEEEpeerreviewmaketitle

\section{Introduction}
\IEEEPARstart{A}{ssisted} navigation of pedestrians and automated driving of intelligent vehicles are inextricably intertwined in the Intelligent Transportation Systems (ITS) field~\cite{cao2020rapid,mancini2018mechatronic,mancini2015embedded,stahlschmidt2015descending,yang2018unifying,xiang2019importance}, both with the aim to improve traffic flow towards the utopia of all road participants. To this end, it is necessary to expand the coverage of assistance systems from drivers to pedestrians, especially those with visual impairments, who are one of the most vulnerable road users~\cite{manduchi2011mobility,Williams-et-al_2013}. 

To assist the navigation of visually impaired people, it is essential to attain efficient and robust \emph{walking} scene understanding, which shares similar challenges with the ITS research line on \emph{driving} surrounding segmentation~\cite{romera2017erfnet,yang2019pass,yang2020omnisupervised}. In real-world scenes, modern fully glazed facades and transparent objects are very common, but they are rarely addressed in existing semantic perception systems either for automated transportation or assisted navigation. Knowledge of glass architecture~\cite{butera2005glass} and glass doors~\cite{maringer2019suitability,mei2020don} are particular important for visually impaired people, because transparent objects often present architectural barriers which hinder the mobility of people with low vision or blindness. For example, if an inaccessible area blocked by a transparent door (Fig.~\ref{fig1:sys}(b)) is detected by an assistive system as accessible, it could lead to a wrong interaction and cause harm to the user. 

However, understanding transparent objects is a puzzle for most vision-based navigation assistance systems~\cite{aladren2014navigation,wang2017enabling,duh2020v}, as notoriously, satisfactory dense prediction of transparent objects is difficult to obtain. 3D vision-based methods hardly recover the depth information of texture-less areas or transparent surfaces~\cite{aladren2014navigation,liu2021hida}, whereas traditional segmentation-based solutions do not cover the transparent object in 2D images~\cite{yang2018unifying,lin2019deep}. Additionally, guide dogs often get confused leading people with blindness to full-pane windows. The appearance difference between glass doors and large glass windows is insignificant, thus it troubles people with residual sight~\cite{saha2019wayfinding}. Moreover, a system to identify landmarks such as doors is particularly appreciated by people with visual impairments, since finding a door or a building entrance is difficult due to the inaccuracy of GPS~\cite{saha2019wayfinding,berenguel2020floor}.

To tackle these challenges, we put forward a wearable system for walking scene perception, covering object- and walkable area segmentation, with the ultimate goal to enable visually impaired people navigate in real-world scenes safely and independently. We propose an efficient transformer-based semantic segmentation architecture dubbed \emph{Trans4Trans}, precisely \emph{Transformer for Transparency}, as shown in Fig.~\ref{fig1:sys}(c). Since transparent objects are often texture-less or share identical content as their surroundings, it is essential to associate long-range visual cues to robustly infer transparent regions.
For this reason, Trans4Trans is established with both transformer-based encoder and decoder to fully exploit the long-range context modeling capacity of self-attention layers in transformers~\cite{vaswani2017attention}.
We design a \emph{Transformer Parsing Module (TPM)} to fuse multi-scale feature maps generated from embeddings of dense partitions. The symmetric transformer-based decoder with a dual-head structure, thereby, is able to consistently parse feature maps from the transformer-based encoder (see Fig.~\ref{fig1:sys}(c)). Along with semantically predicting general thing- and stuff categories like walkable areas, the dual-head design allows to segment transparent objects completely and accurately, which unifies various detection tasks and enhances the perception reliability relevant for safety-critical navigation assistance.

Trans4Trans has been integrated in our wearable assistive system (see Fig.~\ref{fig1:sys}(a)) comprising a pair of smart vision glasses and a mobile GPU processor, which can deliver a holistic scene understanding swiftly and precisely, owing to the high efficiency of our model. Based on the entire semantic information, the user interface is designed with a customized set of acoustic feedback, including the sonification of the detected objects, the identified walkable directions, and warnings of close-range obstacles. The voice user interface yields intuitive navigation suggestions while requiring no prior knowledge.

A comprehensive variety of experiments has been conducted on multiple semantic segmentation datasets~\cite{stanford2d3d,xie2021segmenting} to verify the effectiveness of the assistive system. Particularly, the proposed model achieves state-of-the-art accuracy on the test sets of Stanford2D3D~\cite{stanford2d3d} and Trans10K-v2~\cite{xie2021segmenting}.
Considering the synergy towards traffic safety and the shared challenges between walking- and driving scene understanding, Trans4Trans is further verified on driving scene segmentation benchmarks including Cityscapes~\cite{cityscapes}, ACDC~\cite{sakaridis2021acdc}, and DADA-seg~\cite{zhang2021issafe}, demonstrating its efficiency and robustness for various ITS application scenarios (see Fig.~\ref{fig1:sys}(d)). Finally, a user study with visually impaired people and a series of field tests validate the usability and reliability of our portable system for navigational perception and assistance in the real world. To the best of our knowledge, this is the first work to use vision transformers for assisting people with visual impairments. 

This work is the extension of our conference paper~\cite{zhang2021trans4trans}, extended with a detailed description of the proposed Trans4Trans architecture and an extended set of experiments and analyses. In summary, we deliver the following contributions:
\begin{compactitem}
    \item We build a wearable assistive system with a pair of smart vision glasses and a portable GPU processor
    for aiding people with visual impairments during navigation.
    \item We put forward an efficient segmentation architecture with transformer-based encoder and decoder, namely Transformer for Transparency (\emph{Trans4Trans}). Its dual-head design can unify general- and transparent object segmentation. We propose a Transformer Parsing Module (TPM) to harvest multi-scale feature representations generated from embeddings of dense partitions.
    \item Trans4Trans has surpassed state-of-the-art Convolutional Neural Networks (CNNs) and transformer-based methods on Stanford2D3D and Trans10K-v2 datasets, while maintaining high efficiency on the assistive system.
    \item Trans4Trans is verified on driving scene segmentation benchmarks including Cityscapes, ACDC, and DADA-seg, demonstrating the robustness in different conditions and effectiveness towards real-world applications.
    \item We design an assistive algorithm based on multiple segmentation results and the depth information. We create a user interface via a customized set of acoustic feedback and conduct a user study and various field tests, evidencing the usability and reliability of the assistive system.
\end{compactitem}

\section{Related Work}
\subsection{Semantic Segmentation for Visual Assistance}
Whereas traditional assistance systems rely on multiple monocular detectors and depth sensors~\cite{stahlschmidt2015descending,aladren2014navigation,wang2017enabling,dynamic_crosswalk} for different perception tasks, semantic segmentation allows to solve many navigational detection problems in a unified way and has been employed in visual assistance. As pioneer approaches, semantic paintbrush~\cite{miksik2015semantic} was proposed as an augmented reality system for people with low vision, while semantic labeling was applied to the problem of navigation using prosthetic vision~\cite{horne2016semantic}.

Yang~\etal~\cite{yang2018unifying} put forward to seize pixel-wise semantic segmentation to unify terrain detection tasks covering traversability perception, stairs navigation, and water-hazards negotiation. Mao~\etal~\cite{mao2021panoptic} argued for efficient panoptic segmentation towards a comprehensive labeling and verified the performance in adverse conditions like snowy scenes. In~\cite{long2019unifying,yohannes2019content}, instance-specific segmentation models like Mask R-CNN~\cite{he2017mask} were directly used for content-aware sensing. Liu~\etal~\cite{liu2021hida} proposed HIDA to enable holistic indoor understanding for detection and avoidance of obstacles based on 3D point cloud semantic instance segmentation. Dense semantic segmentation has also been leveraged to address intersection perception like the detection of crosswalks, sidewalks, and blind roads~\cite{cao2020rapid,yang2018intersection,hsieh2020outdoor}.
Moreover, we observe the trend of using semantic segmentation in various navigation assistance platforms~\cite{duh2020v,lin2019deep,mahendran2021computer}. Yet, both of these systems cannot resolve the problems emerged with transparent objects.

\subsection{Transparent Object Sensing}
Classical visual assistance systems~\cite{bai2017smart,huang2018glass} resorted to multi-sensor fusion to overcome the difficulties in handling transparent obstacles like glass objects, French windows and French doors by using ultrasonic sensors together with RGB-D cameras.
Multimodal and multispectral information were also frequently used. Chen~\etal~\cite{chen2018improving} enhanced the depth estimation via consumer RGB-D cameras by fusing infrared stereo vision and color stereo vision.
Okazawa~\etal~\cite{okazawa2019simultaneous} performed simultaneous recognition of ordinary non-transparent objects and transparent objects by utilizing the difference in transmission characteristics under multispectral scenes. Moreover, polarization cues~\cite{deep_polarization_cues} and reflection priors~\cite{lin2021rich} were often explored for transparency perception. For instance, Xiang~\etal~\cite{xiang2021polarization} built a polarization-driven semantic segmentation architecture by bridging RGB and polarization dimensions dynamically using efficient attention connections, which considers the optical features of polarimetric information for robust representation of diverse materials and lifts the performance of classes with polarization properties like \emph{glass}.

Recently, large-scale transparent object segmentation datasets emerge~\cite{mei2020don,xie2021segmenting,okazawa2019simultaneous,xiang2021polarization,xie2020segmenting}. Mei~\etal~\cite{mei2020don} constructed the glass detection dataset in daily-life scenes. Xie~\etal~\cite{xie2021segmenting,xie2020segmenting} built the Trans10K dataset and validated that while pure RGB-based transparent object segmentation is rather a largely unsolved task, it is potential for real-world usages with the increased data amount. This spurs the community to go beyond traditional perception paradigms relying on sensor fusion and develop novel methods addressing transparent object segmentation. AdaptiveASPP~\cite{fakemix} was designed as an enhanced version of ASPP~\cite{deeplabv3+} to appropriately harvest rich features at multi-stage levels in joint segmentation and boundary predictions. EBLNet~\cite{enhanced_boundary_learning} integrated a point-based graph convolution module to model global shape representations. Whereas these methods have reached high accuracy on glass-like object detection, we aim for a both efficient and robust semantic segmentation desirable for real-world navigation assistance. We establish a transformer-based system to assist transparency perception.

\subsection{Attention- and Transformer-based Semantic Segmentation}
Since Fully Convolutional Networks (FCNs)~\cite{fcn} achieved semantic segmentation end-to-end by viewing it as a dense-pixel classification task, modern methods develop upon this paradigm by augmenting FCNs with context aggregation modules. PPM~\cite{pspnet} uses multiple scales of pooling operators in a pyramid manner, while ASPP~\cite{deeplabv3+} leverages atrous convolutions of different dilation rates to enlarge receptive fields. DANet~\cite{danet}, OCNet~\cite{ocnet}, and CCNet~\cite{ccnet} devise variants of non-local attention blocks to exploit long-range pixel relations. Disentangled and asymmetric versions~\cite{yin2020disentangled,zhu2019asymmetric,yang2021capturing} of non-local modules have also been designed to reduce computation complexity of dense pixel-pair associations.

Inspired by Vision Transformer~\cite{vit} that utilizes transformer layers to sequences of image patches for visual recognition, SETR~\cite{setr} and Segmenter~\cite{segmenter} directly append upsampling and segmentation heads atop ViT, encoding long-range context information from the very first layer. Leveraging the advance of DETR~\cite{detr}, MaX-DeepLab~\cite{maxdeeplab} and MaskFormer~\cite{maskformer} look at image segmentation from the lens of mask prediction and classification. With these successes, various transformer architectures for dense image segmentation appear~\cite{swin,segformer,fully_transformer_networks}. PVT~\cite{pvt,wang2021pvtv2} and SegFormer~\cite{segformer} propose pyramid structures of vision transformers for collecting hierarchical feature representations. ECANet~\cite{yang2021capturing} and CSWin transformer~\cite{dong2021cswin} advocate performing self-attention in horizontal or vertical stripes to achieve powerful modeling capacity while lowering computation overheads.

In this research, we propose an efficient \emph{Trans4Trans} framework with focus set on assisting navigation of visually impaired people in the wild. Differing from existing works that either stack attention layers~\cite{danet,yang2021capturing} and encoder-decoder transformers on CNN backbones~\cite{xie2021segmenting}, or employing CNN-based decoders on top of transformer encoders~\cite{setr,pvt}, in \emph{Trans4Trans} both encoder and decoder are based on transformer, together with a novel Transformer Parsing Module inserted in the dual-head decoder, which unifies transparent object and semantic scene segmentation.



\begin{figure*}[!t]
    \centering
    \includegraphics[width=0.99\linewidth]{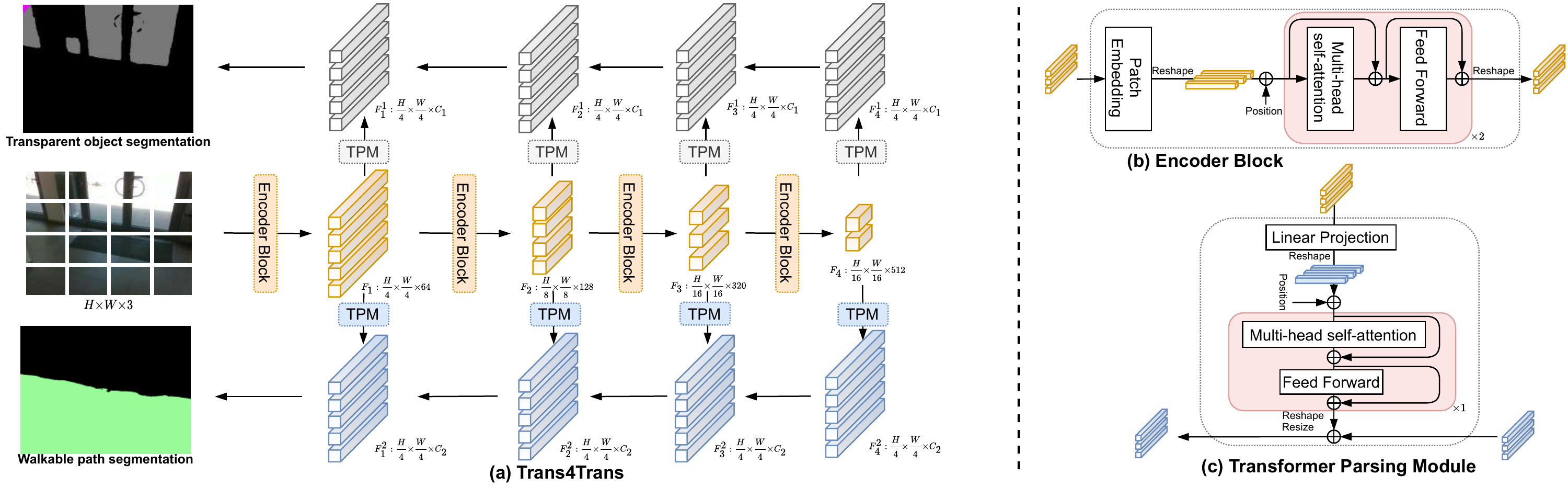}
    \caption{\small The architecture of (a) Trans4Trans model consists of shared encoder and dual decoders, while (b) and (c) are the general transformer-based encoder block and our proposed Transformer Parsing Module~(TPM) for decoder, respectively.}
    \label{fig:arch_trans4trans}
\end{figure*}

\section{System Architecture}

\subsection{Trans4Trans}
Transformers adapted to vision tasks have achieved competitive performance compared to the CNN counterparts in recent years. Benefiting from the Multi-Head Self-Attention (MHSA) structure, ViT~\cite{vit} introduced the modeling ability in acquiring long-range dependencies, which is vital for dense prediction tasks, \eg transparency- and semantic perception. SETR~\cite{setr} extended vision transformer to segmentation and validated the feasibility of performing MHSA on high-resolution inputs. In order to reduce the costly computation of full-resolution MHSA, Swin~\cite{swin} and CSWin~\cite{dong2021cswin} limited the token region of the full attention by using shift operations or cross-shaped windows. However, the global context is slowly covered by stacking a large number of such regional self-attention blocks, and the receptive field is rather limited. In this work, we preserve the full-scale self-attention in the encoder for faster-enlarging receptive fields. To balance inference efficiency and segmentation accuracy in Trans4Trans, we consider: (1) to build upon a pyramidal transformer encoder with full-scale self-attention blocks; (2) to lighten the decoder to enable deploying a dual-head model on portable GPUs; and (3) to maintain a symmetrical encoder-decoder structure for obtaining pyramidal feature representations.

\noindent \textbf{Pyramidal Feature Representations.}
The whole architecture of the Trans4Trans model is shown in Fig.~\ref{fig:arch_trans4trans}(a). 
Similar to recently emerged models~\cite{segformer}\cite{dong2021cswin}, we split the given image $H{\times}W{\times}3$ with a $4{\times}4$ patch size, as a smaller patch size is more conducive for dense predictions. A $7{\times}7$ convolution layer with a stride of $4$ is utilized to perform overlapping patch embedding. 
Contrary to the single-scale feature of ViT, pyramidal features $\{F_1, F_2, F_3, F_4\}$ are obtained in our hierarchical encoder. 
Similar to the traditional ResNet~\cite{resnet} model, the whole encoder is composed of four stages, within which a $3{\times}3$ convolution layer with a stride of $2$ is utilized to progressively downsample the lower-level high-resolution features to the higher-level low-resolution features. 
In each stage of the efficient Trans4Trans model, two full-scale MHSA blocks are stacked to realize faster-enlarging receptive fields. The encoder block is shown in Fig.~\ref{fig:arch_trans4trans}(b), in which we use depth-wise convolutions in the Feed Forward module to reduce the computation.
In specific, the pyramid features are extracted with $\{4, 8, 16, 32\}$ downsampling rates and $\{64, 128, 320, 512\}$ channels.

\noindent \textbf{Transformer Parsing Module.}
Recent Transformers~\cite{setr}\cite{swin}\cite{twins} opt to modify the encoder while neglecting the decoder design. Yet, a robust decoder is critical for deploying a transformer model in real applications, especially for an assistive system. The whole decoder is composed of four transformer-based stages corresponding to the encoder, which is different to the single-scale decoder of ViT~\cite{vit} and CNN-headed decoder of Trans2Seg~\cite{xie2021segmenting}. To maintain the symmetrical pyramid features both in the encoder and the decoder, we propose the \emph{Transformer Parsing Module~(TPM)} to interpret the multi-scale features $\{F_1, F_2, F_3, F_4\}$, as illustrated in Fig.~\ref{fig:arch_trans4trans}(a) and TPM is shown in detail in Fig.~\ref{fig:arch_trans4trans}(c). A linear projection operation is used to translate the feature from the encoder into a preset embedding with a dimension $C$, which maintains the features channel-wise identical throughout the decoder. An MHSA module similar to the one in the encoder block is leveraged to obtain symmetrical features. An upsampling (resize) operation is used to preserve the same resolution $\frac{H}{4}\times{\frac{W}{4}}\times{C}$ of different features between adjacent stages. As a result, the lower-resolution features with long-range contextual information are aggregated with the higher-resolution features with fine and local information in the lightweight hierarchical decoder.

\noindent \textbf{Multiple Decoders.}
Based on TPM, a full-attended and efficient decoder demands little computing resources and eases the integration with arbitrary encoder backbones, thus a segmentation transformer model is more flexible to be deployed on a portable hardware system. However, training a robust transformer model requires a large-scale dataset~\cite{vit}. To deal with the data-hunger problem and robustify segmentation in the wild, we design a double-head model to perform joint training on multiple datasets, \eg general and transparent object segmentation datasets for our assistive system, or normal and adverse driving scene segmentation datasets for automated vehicles. The feature maps representation in two decoder heads are illustrated as $\{F_1^1, F_2^1, F_3^1, F_4^1\}$ and $\{F_1^2, F_2^2, F_3^2, F_4^2\}$ in Fig.~\ref{fig:arch_trans4trans}(a).
Benefiting from our proposed TPM, the amount of GFLOPs and parameters of this dual-head structure is largely reduced compared to deploying two separate models. Also importantly, diverse features can be learned from various datasets.
For real-world semantic segmentation, mounting multiple decoder heads robustifies the feature learned via the shared encoder and prevents overfitting. Meanwhile, the entire model is computationally efficient.

\subsection{Portable Assistive System}

The entire wearable navigation assistance system is composed of a pair of smart vision glasses and a mobile GPU processor, \eg, a lightweight laptop or an NVIDIA AGX Xavier (Fig.~\ref{fig1:sys}).

\begin{figure}[t]
    \centering
    \includegraphics[width=0.99\linewidth]{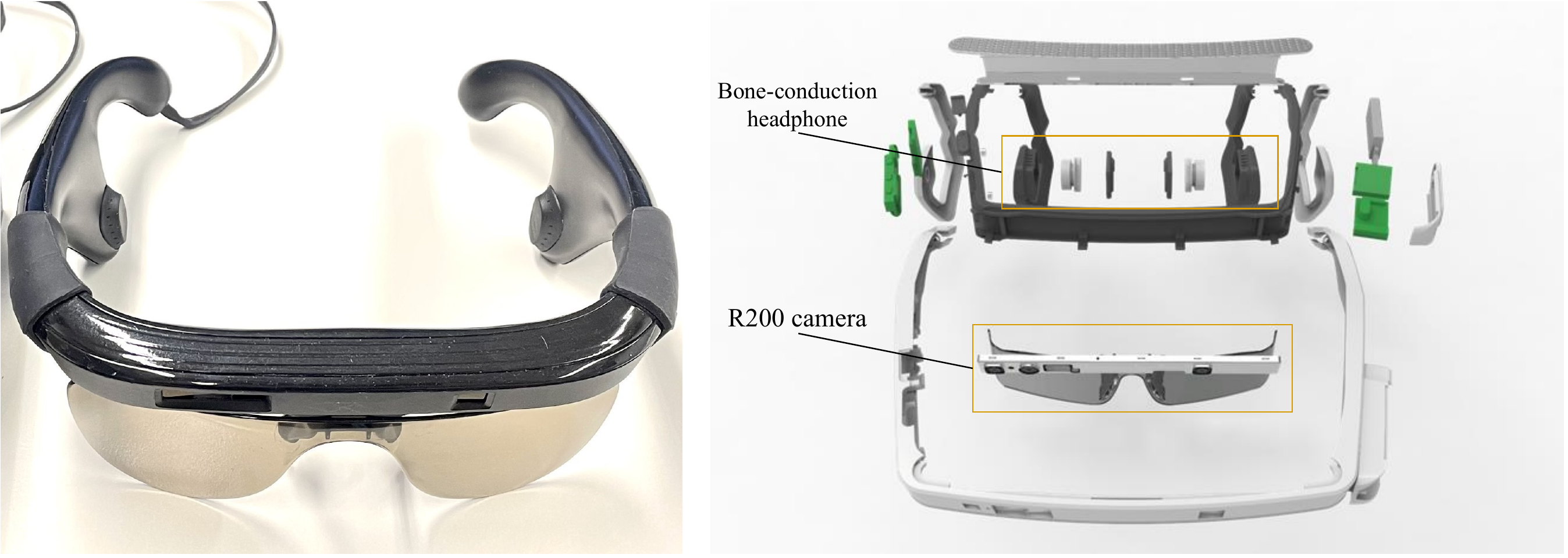}
    \caption{\small Detailed illustration of components in the smart vision glasses designed for assisting visually impaired people. The main components: RealSense R200 camera and bone-conduction headphones.}
    \label{fig:glasses}
\end{figure}

As shown in Fig.~\ref{fig:glasses}, the smart vision glasses have been integrated with a RealSense R200~\cite{keselman2017intel} RGB-D sensor to enable real-time acquisition of RGB and depth images at the resolution of $640{\times}480$, and a pair of bone-conduction earphones for delivering acoustic feedback to people with visual impairments. This is crucial as visually impaired people often rely on the sounds from the surroundings for determining the orientation and bone-conduction headphones will not block their ears when using the assistive system. 

The equipped RealSense R200 sensor uses a combination of active speckle projecting and passive stereo matching. With this design, it can work well both indoors and outdoors. In texture-less indoor scenes, the projected infrared speckles will augment the environments, which are beneficial for stereo matching algorithms (\eg, R200 leverages a straightforward correlation engine~\cite{keselman2017intel}) to yield dense depth estimation. In sunny outdoor scenes, although many projected patterns would be overwhelmed by sunlight, the infrared components of natural light shine on the scene to form well-textured infrared image pairs, thus enabling robust depth sensing. In our assistive system, depth information is mainly used to assist the obstacle avoidance function, \eg, to prioritize near-range objects over mid- and long-range objects.

\subsection{System Algorithm}\label{sec:sys_algo}
The algorithm and user interface of our assistive system with the dual-head Trans4Trans are described in Algorithm~\ref{algo:system}.

\noindent\textbf{System setting.}
According to the properties of the R200 camera sensor~\cite{keselman2017intel}, the frame rate of RGB-D stream can be set up to $60$. When the camera frame rate is set to $30$ and the resolution is set to $640{\times}480$, the minimum effective distance of depth information is around $1.5m$. However, it cannot cover the perception of objects in close range ($<1.5m$). To guarantee a timely data acquisition and widen the minimum effective range of depth information to $0.5m$, we preset the frame rate of R200 to $60$ and the resolution to $320{\times}240$.
Based on this setting, the camera can obtain sufficiently-accurate depth information and cover near-range objects, which is critical for the assistive system.
Once the system starts, it repeats the whole algorithm every $n$ seconds. 
According to our experiments, the time interval setting as $1$--$2$ seconds effectively prevents cognitive overload, especially in cases of complex scenes containing a large number of objects. Still, it is adjustable depending on the demands of users, \eg, a short interval for more feedback to explore unknown space. Within $2$ seconds, our efficient Trans4Trans model can perform segmentation of approximately $20$ frames. After aggregating multiple segmentation results, the final feedback of the system is more reliable against noises and perturbations.

\begin{algorithm}[!t]
	\caption{Assistive system}
	\label{algo:system}
	\KwData{RGB-D as ${X}\in \mathcal{R}^{H \times W \times 3}$ and ${D}\in \mathcal{R}^{H \times W}$.}
	\KwResult{General segmentation $G\in \mathcal{R}^{{H}\times{W}\times{13}}$; Transparency $T \in \mathcal{R}^{{H}\times{W}\times{11}}$; }
    initialize walkable rate: $R_{l}, R_{f}, R_{r}$, parameters: $\theta_{obstacle}$, $\theta_{trans}$, $\theta_{walkable}$ \;
    \While{system start and each $n$ seconds }{
    RGB-D update and Trans4Trans segmentation: \\
    $G_{path}\in \mathcal{R}^{H \times W}, G_{object}\in \mathcal{R}^{H \times W \times 12}$ \;
    ${T_{stuff}\in \mathcal{R}^{H \times W \times 3}, T_{thing}\in \mathcal{R}^{H \times W \times 8}}$ \;
    partition $\{R_{l}, R_{f}, R_{r}\} \leftarrow G_{path}$ \;
    \uIf{$\overline{D} < \theta_{obstacle}$}{
    $vibration$ as obstacle warning\;
    }
    \uElseIf{$max\{\overline{T}_i\}\in T_{stuff} > \theta_{trans}$}{
    $speech \leftarrow argmax\{\overline{T}_i\}\in T_{stuff}$ ;
    }
    \uElseIf{$max\{R_{l}, R_{f}, R_{r}\} > \theta_{walkable}$}{
    $speech \leftarrow argmax\{R_{l}, R_{f}, R_{r}\} \in \{left, forward, right\}$;
    }
    \Else{
    $speech \leftarrow nearest\{T_{thing}, G_{object}\}$\;
    }
    }
\end{algorithm}

\noindent\textbf{Obstacle avoidance.}
When moving indoors with limited space, the building materials and densely arranged objects will seriously hinder the flexibility of using the common white cane as an obstacle avoidance aid.
In order to tackle the collision issue and balance indoor and outdoor scenarios, our system presets the highest priority for obstacle avoidance.
In contrast to the fixed and limited categories defined in an obstacle avoidance engine~\cite{lin2019deep}, we leave the obstacles as open-set and detect them based on the depth information ${D}\in\mathcal{R}^{H{\times}W}$.
In other words, if the average value of the depth information $\overline{D}$ is smaller than the preset distance threshold $\theta_{obstacle}$, the user will be immediately notified in the form of \emph{vibration}. To minimize the uncertainty of vibrations and the cognitive load, only one single default threshold is set, instead of setting various vibration frequencies for different distances. Another purpose is to preclude the chaotic and low-confidence segmentation from the less-textured images when users walk too close and face to the object surface, such as images of white wall or doors. According to our pre-tests and the minimum effective range ($0.5m$) of the R200 camera, we set the distance threshold as $\theta_{obstacle}=1m$, and thereby the system can effectively detect open-set and near-range obstacles and output \emph{vibration} as warnings.

\noindent\textbf{(Transparent) object segmentation.}
After receiving the RGB image ${X}\in\mathcal{R}^{H{\times}W{\times}3}$, our efficient dual-head Trans4Trans model outputs two segmentation predictions, which are general object segmentation $G\in\mathcal{R}^{H{\times}W{\times}{13}}$ and transparent object segmentation $T\in\mathcal{R}^{H{\times}W{\times}{11}}$, respectively. The specific object categories of each segmentation result will be introduced later in the dataset description. The general object segmentation is divided into $G_{path}$ of \emph{walkable path}, \ie, \emph{floor} class, and $G_{object}$ of other \emph{objects}~\cite{stanford2d3d}.
Besides, the transparent object segmentation is divided into two disjoint sets as: $T_{stuff}\in\mathcal{R}^{H{\times}W{\times}3}$ with $\{$\emph{window, glass door, glass wall}$\}$, and $T_{things}\in \mathcal{R}^{H{\times}W{\times}8}$ with $\{$\emph{shelf, jar/tank, freezer, eyeglass, cup, bowl, bottle, box}$\}$~\cite{xie2021segmenting}. 
To combine the segmentation results $G_{path}$ and $T_{stuff}$, we tend to preset a higher priority of prompts for the transparent objects. Specifically, when the maximum segmented area of transparent stuff is greater than a preset threshold $\theta_{trans}$, its corresponding category is fed back in a speech form, before the one of other general objects. Based on our experiments, when the whole image area is $1.0$, we set the $\theta_{trans}=0.5$ to eliminate the effects of jitter-errors in segmentation.

\noindent\textbf{Walkable path detection.}
After achieving general object segmentation, the walkable mask $G_{path}$ is further partitioned into three regions as $\{left, forward, right\}$ directions for orientation assistance. The local ratio $R$ is calculated by $ G_{path}^i / A^i_{image}$, where $i$ represents one of the three directions and $A$ denotes the image area. As a result, the horizontally divided ratios are denoted as $\{R_{l}, R_{f}, R_{r}\} \leftarrow G_{path}$.
Then, an intuitive and effective strategy is to prompt the direction that has the largest walkable area, only when the largest local ratio $R$ is greater than the preset threshold $\theta_{walkable}$. In this case, we set a strict threshold $\theta_{walkable}=0.4$ to ensure that the maximum area is safe enough and walkable for the user. According to our test, this orientation approach guarantees anti-veering in a straight path outdoors and indoors. Furthermore, it can accurately predict the best instantaneous turning direction during walking at an intersection, so as to constantly yield a safe direction suggestion.

\section{Experiments}
In this section, with extensive experiments on multiple datasets, we verify the efficiency and robustness of the proposed Trans4Trans architecture for general- and transparent object segmentation, as well as driving scene segmentation.
We first describe the datasets and settings in Sec.~\ref{sec:settings}. Then, we quantitatively measure the accuracy of general- and transparent object segmentation desired for real-world navigation assistance (Sec.~\ref{sec:accuracy_transparent}) as well as the robustness of driving scene segmentation in diverse conditions (Sec.~\ref{sec:driving}). In Sec.~\ref{sec:realtime}, we assess the real-time performance and in Sec.~\ref{sec:feature}, we investigate the importance of context priors for transparency perception with feature visualizations. Finally, in Sec.~\ref{sec:qualitative}, we analyze qualitative segmentation results.

\subsection{Datasets and Settings}
\label{sec:settings}
\noindent\textbf{Trans10K-v2}~\cite{xie2021segmenting} has $10,428$ images, which are divided into $5,000$, $1,000$, and $4,428$ for training, validation, and testing, respectively. The $H{\times}W$ resolution of the images is $835{\times}1,113$. There are $11$ object categories marked as \emph{shelf, jar or tank, freezer, window, glass door, eyeglass, cup, wall, glass bow, water bottle}, and \emph{storage box}.

\noindent\textbf{Stanford2D3D}~\cite{stanford2d3d} has $70,496$ images. The resolution of the images is $1,080{\times}1,080$. The $13$ object categories are annotated as \emph{beam, board, bookcase, ceiling, chair, clutter, column, door, floor, sofa, table, wall}, and \emph{window}. Based on the fold-1 data splitting~\cite{stanford2d3d}, the dataset is divided into the training set with \emph{Area 1-4} and \emph{Area 6}, the validation set with \emph{Area 5a}, and the test set with \emph{Area 5b}. There are $52,905$, $6,261$, and $11,332$ images in the training-, validation-, and test set, respectively. 

\noindent\textbf{Cityscapes}~\cite{cityscapes} is a street scene dataset captured in $50$ different European cities under \emph{normal} conditions. The dataset comprises $2,979$ and $500$ images for training and validation. The resolution of the images is $2,048{\times}1,024$ and the images are annotated with $19$ categories.

\noindent\textbf{ACDC}~\cite{sakaridis2021acdc} is a driving scene dataset captured under four \emph{adverse} conditions, \ie, \emph{fog, night, rain}, and \emph{snow}. It contains $1,600$ training- and $406$ validation images which are publicly available, and $2,000$ test images for benchmarking. The resolution of the images is $1,920\times1,080$.

\noindent\textbf{DADA-seg}~\cite{zhang2021issafe} is a testing dataset captured in traffic \emph{accidental} scenes. The resolution of image is $1,584\times660$. It contains $313$ images densely annotated with $19$ categories, which are consistent with Cityscapes for benchmarking semantic segmentation. These extreme accident scenarios are utilized in our work together with ACDC to study the robustness of Trans4Trans in different adverse conditions.

\noindent\textbf{Implementation details.}
Our Trans4Trans model is implemented with  CUDA 11.2 and PyTorch 1.8.0 with an initialized learning rate $1e-4$ and scheduled by the poly strategy~\cite{bisenet} with power $0.9$ in $100$ epochs. AdamW~\cite{adam_optimization} is chosen as the optimizer with epsilon $1e-8$ and weight decay $1e-4$ and batch size is set as $4$ on each of four 1080Ti GPUs. The experiments for ablating the effect of embedding channels are conducted on a single GPU. The images in the training and testing stages are resized in the resolution of $512{\times}512$ or $768{\times}768$ (will be specified) for the experiments, to maintain the shape of position embedding. For a fair comparison with the previous state-of-the-art Trans2Seg~\cite{xie2021segmenting}, some tricks such as OHEM, auxiliary and class-weighted losses are not applied in the experiments. We use mean Intersection over Union (mIoU) as the main evaluation metric.

\subsection{Accuracy of General and Transparent Object Segmentation}
\label{sec:accuracy_transparent}
In this subsection, we conduct analysis of the performance of Trans4Trans on different datasets and the performance of varied encoder-decoder structures realized by different combinations of CNN/transformer in detail to verify the efficacy of our approach. Then, computation complexity in GFLOPs of different approaches and segmentation accuracy are presented and compared with state-of-the-art methods.

\begin{table}[!t]
\centering
\resizebox{\columnwidth}{!}{
\begin{tabular}{@{}lll|cc|ll@{}}
\toprule
\textbf{Network} & \textbf{Encoder} & \textbf{Decoder} & \textbf{GFLOPs} & \textbf{MParams} & \textbf{Stanford2D3D} & \textbf{Trans10K-v2}\\ \midrule \midrule
Trans2Seg-T~\cite{xie2021segmenting}   & PVT-T~\cite{pvt} & \multirow{3}{*}{Transformer~\cite{xie2021segmenting}} &10.16  &13.11 & 41.00 & 64.60    \\
Trans2Seg-S~\cite{xie2021segmenting}   & PVT-S~\cite{pvt} &   &19.58  &24.36 & 41.89 & 68.47     \\
Trans2Seg-M~\cite{xie2021segmenting}   & PVT-M~\cite{pvt} &   &49.00  &56.20 & 42.49 & 72.10     \\ 
\midrule

Trans4Trans-T & PVT-T~\cite{pvt}    & \multirow{3}{*}{TPM / Single-head} &10.45  &12.71 & 41.28\gbf{+0.28} & 68.63\gbf{+4.03}    \\
Trans4Trans-S & PVT-S~\cite{pvt}    &  &19.92  &23.95 & 44.47\gbf{+3.04} & 74.15\gbf{+5.68}     \\
Trans4Trans-M & PVT-M~\cite{pvt}    &  &34.38  &43.65 & 45.73\gbf{+3.24} & 75.14\gbf{+3.04}     \\
\midrule
Trans4Trans-T & PVT-T~\cite{pvt}    & \multirow{3}{*}{TPM / Dual-head} &  11.22 & 13.10 & 40.44\rbf{-0.56} & 69.84\gbf{+5.24}   \\
Trans4Trans-S & PVT-S~\cite{pvt}    &  &  20.69 & 24.34 & 43.45\gbf{+1.56} & 74.57\gbf{+6.10}     \\
Trans4Trans-M & PVT-M~\cite{pvt}    &  &  35.17 & 44.04 & 45.15\gbf{+2.66} & 74.98\gbf{+2.88}   \\
\bottomrule[1px]
\end{tabular}
}
\caption{\small Comparison with state-of-the-art methods on Stanford2D3D~\cite{stanford2d3d} and Trans10K-v2~\cite{xie2021segmenting}. 
\#MParams, \#GFLOPs are calculated with the input size of $512{\times}512$. The embedding dimension of Trans4Trans is $64$, while it is $\{128,128,256\}$ for Trans2Seg-T/-S/-M decoder,  following~\cite{xie2021segmenting}.
}
\label{tab:performance}
\end{table}

\noindent\textbf{Results.}
The performances of different encoder-decoder model architectures are shown in Table~\ref{tab:performance}. Compared with Trans2Seg~\cite{xie2021segmenting}, the encoder and decoder in our approach are both transformer-based. The single- and dual-head decoders are constructed with the TPM design. As shown in Table~\ref{tab:performance}, the single-head Trans4Trans clearly outperforms Trans2Seg in all settings (T, S, and M are short for Tiny, Small, and Medium). Trans4Trans-M achieves the best performance in mIoU with $45.73\%$ on Stanford2D3D dataset and $75.14\%$ on Trans10K-v2 dataset, exceeding by more than $3\%$ \wrt Trans2Seg-M on the challenging transparent object segmentation benchmark. Meanwhile, compared to Trans2Seg-M, the computational complexity in GFLOPs is much smaller in our approach Trans4Trans-M, while Trans4Trans-Tiny and Trans4Trans-Small also achieve better performance than the corresponding Trans2Seg variant on Trans10K-v2.

\begin{table}[!t]
\small
\centering
\resizebox{\columnwidth}{!}{
\begin{tabular}{c|cc|cc|cc}
\toprule
\textbf{Method} & \textbf{Trans. Enc.} & \textbf{CNN Enc.} & \textbf{Trans. Dec.} & \textbf{CNN Dec.} & \textbf{GFLOPs}$\downarrow$ & \textbf{mIoU}~(\%)~$\uparrow$ \\ 
\midrule
FCN~\cite{fcn}                      &                &  \checkmark   &               & \checkmark   & 42.2  & 62.7 \\ 
OCNet~\cite{ocnet}                  &                &  \checkmark   &               & \checkmark   & 43.3 & 66.3 \\ 
Trans2Seg~\cite{xie2021segmenting}  &                &  \checkmark   & \checkmark    &              & 40.9 & 69.2 \\ 
PVT~\cite{pvt}                      & \checkmark     &               & \checkmark    &              & 49.0 & 72.1 \\
Trans4Trans (ours)                  & \checkmark     &               & \checkmark    &              & \textbf{34.3} & \textbf{75.1} \\
\bottomrule
\end{tabular}
}
\caption{\small Effect of CNN/Transformer combination.
Models are evaluated on the Trans10K-v2 dataset. \#GFLOPs are calculated with the input size of $512{\times}512$.}
\label{tab:combination}
\end{table}

When incorporating more general knowledge by learning jointly with supervision from Stanford2D3D dataset, dual-head Trans4Trans consistently improves the performance on Trans10K-v2 compared with Trans2Seg. More importantly, dual-head Trans4Trans highly alleviates the problem of overfitting and reduces false positives of transparent obstacle warning observed in our field tests, and thereby it is more suitable for real-world navigation perception. Overall, the superiority and efficiency of Trans4Trans are verified through the experiments on transparent- and general object segmentation.

\noindent\textbf{Combination of CNN / Transformer.}
As shown in Table~\ref{tab:combination}, varied combinations of CNN-/Transformer-based encoder and decoder are compared, where FCN~\cite{fcn} and OCNet~\cite{ocnet} are composed of only CNNs, whereas Trans2Seg is composed of CNN-based encoder and transformer-based decoder. In contrast, the architecture in the proposed Trans4Trans approach is a fully transformer-based encoder-decoder model. Our proposed approach outperforms both these aforementioned competitive networks such as OCNet and transformer-based encoder-decoder architectures such as PVT. In addition, our Trans4Trans keeps clearly smaller GFLOPs while being more accurate, demonstrating its suitability for efficient transparent object segmentation.

\begin{figure}[t]
    \centering
    \includegraphics[width=0.99\linewidth]{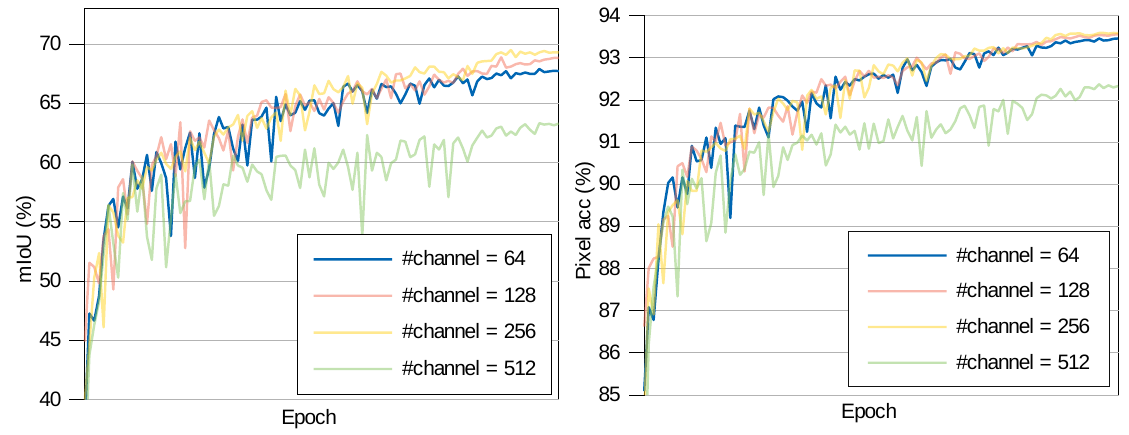}
    \caption{\small mIoU and pixel accuracy curves under different numbers of embedding channels in TPM.}
    \label{fig:channel_test_miou}
\end{figure}

\noindent\textbf{Ablation of TPM channel.}
As TPM is one of our critical designs, the ablation study of different numbers of embedding channels applied in the Trans4Trans decoder is conducted, where the effects are illustrated in Table~\ref{tab:channel_decoder} and Fig.~\ref{fig:channel_test_miou}. It reveals that the larger channel number, the better performance, until $256$. The drop at $512$ indicates that the decoder overfits the encoded feature as shown in Fig.~\ref{fig:channel_test_miou} and the computation complexity becomes exceedingly large with the increase of channel number for TPM. For the response time-critical wearable system, we adopt the smallest $64$ channels when deploying dual-head Trans4Trans due to its highest efficiency and good segmentation performance.
Yet, to pursue high accuracy on driving-scene datasets, we set TPM channel of Trans4Trans-T/-S/-M as $\{64,128,256\}$, respectively.

\begin{table}[h]
\centering
\resizebox{.85\columnwidth}{!}{
\begin{tabular}{c|cccc}
    \toprule
    {\textbf{Channel}}&{\textbf{MParams}}&{\textbf{GFLOPs}}&\textbf{Acc~(\%)}&\textbf{mIoU~(\%)}\\
    \midrule
    64     & 10.45 & 12.71 & 93.46 & 67.89 \\
    128    & 12.46 & 14.02 & 93.49 & 68.88 \\
    256    & 20.41 & 17.82 & 93.58 & 69.51 \\
    512    & 51.50 & 33.82 & 92.37 & 63.33 \\
    \bottomrule
\end{tabular}}
\caption{\small Effect of embedding channel in TPM. All tiny Trans4Trans are trained on Trans10K-v2 at $512{\times}512$ on one GPU. Acc denotes pixel accuracy.}
\label{tab:channel_decoder}
\end{table}

\begin{table*}[!t]
\centering
\scalebox{0.85}{
\begin{tabular}{p{85pt}<{\raggedright}p{30pt}<{\centering}p{22pt}<{\centering}p{22pt}<{\centering}p{30pt}<{\centering}p{18pt}<{\centering}p{22pt}<{\centering}p{18pt}<{\centering}p{18pt}<{\centering}p{18pt}<{\centering}p{18pt}<{\centering}p{18pt}<{\centering}p{18pt}<{\centering}p{18pt}<{\centering}p{18pt}<{\centering}p{18pt}<{\centering}}
\toprule
\multirow{2}{*}{\textbf{Method}} & \multirow{2}{*}{\textbf{GFLOPs}~$\downarrow$} & \multirow{2}{*}{\textbf{ACC}~$\uparrow$} & \multirow{2}{*}{\textbf{mIoU}~$\uparrow$} & \multicolumn{12}{c}{\rule{0pt}{10pt}\textbf{Category IoU}~$\uparrow$} \\ \cline{5-16} 

\rule{0pt}{10pt} &  \multicolumn{1}{c}{} &  &  \multicolumn{1}{c}{} & Background & Shelf  & Jar/Tank  & Freezer & Window & Door & Eyeglass & Cup & Wall & Bowl & Bottle & Box  \\ \midrule
 
\rule{0pt}{10pt} FPENet~\cite{fpenet}  &\multicolumn{1}{|c|}{0.76} & 70.31 & \multicolumn{1}{c|}{10.14} & 74.97 & 0.01 & 0.00 & 0.02 & 2.11 & 2.83 & 0.00 & 16.84 & 24.81 & 0.00 & 0.04 & 0.00 \\[-1.0ex] 
\rule{0pt}{10pt} ESPNetv2~\cite{espnetv2}  & \multicolumn{1}{|c|}{0.83} & 73.03 & \multicolumn{1}{c|}{12.27} & 78.98 & 0.00 & 0.00 & 0.00 & 0.00 & 6.17 & 0.00 & 30.65 & 37.03 & 0.00 & 0.00 & 0.00 \\[-1.0ex] 
\rule{0pt}{10pt} ContextNet~\cite{contextnet} & \multicolumn{1}{|c|}{0.87} & 86.75 & \multicolumn{1}{c|}{46.69}   & 89.86 & 23.22 & 34.88 & 32.34 & 44.24 & 42.25 & 50.36 &  65.23 & 60.00 & 43.88 & 53.81 & 20.17 \\[-1.0ex] 
\rule{0pt}{10pt} FastSCNN~\cite{fastscnn} & \multicolumn{1}{|c|}{1.01} & 88.05 & \multicolumn{1}{c|}{51.93} & 90.64 & 32.76 & 41.12 & 47.28 &  47.47 & 44.64 & 48.99 & 67.88 & 63.80 & 55.08 & 58.86 & 24.65 \\[-1.0ex] 
\rule{0pt}{10pt} DFANet~\cite{dfanet}  & \multicolumn{1}{|c|}{1.02} & 85.15 & \multicolumn{1}{c|}{42.54} & 88.49 & 26.65 & 27.84 & 28.94 & 46.27 & 39.47 & 33.06 & 58.87 & 59.45 & 43.22 & 44.87 & 13.37 \\[-1.0ex] 
\rule{0pt}{10pt} ENet~\cite{enet}  & \multicolumn{1}{|c|}{2.09} & 71.67 & \multicolumn{1}{c|}{8.50} & 79.74 & 0.00 &  0.00 & 0.00 & 0.00 & 0.00 & 0.00 & 0.00 & 22.25 & 0.00 & 0.00 & 0.00 \\[-1.0ex] 
\rule{0pt}{10pt} DeepLabv3+MBv2~\cite{mobilenetv2} & \multicolumn{1}{|c|}{2.62} & 88.39 & \multicolumn{1}{c|}{54.16} & 89.95 & 31.79  & 48.29  & 46.18 & 41.39 & 43.42 & 61.97 & 69.48 & 61.65 & 54.89 & 63.47 & 37.36  \\[-1.0ex] 
\rule{0pt}{10pt} HRNet\_w18~\cite{hrnet}  & \multicolumn{1}{|c|}{4.20} & 89.58 & \multicolumn{1}{c|}{54.25} & 92.47 & 27.66 & 45.08 & 40.53 & 45.66 & 45.00 & 68.05 & 73.24 & 64.86& 52.85 & 62.52 & 33.02   \\[-1.0ex] 
\rule{0pt}{10pt} HarDNet~\cite{hardnet}  & \multicolumn{1}{|c|}{4.42} & 90.19 & \multicolumn{1}{c|}{56.19} & 92.87 & 34.62 & 47.50 & 42.40 & 49.78 & 49.19 & 62.33 & 72.93 & 68.32 & 58.14 & 65.33 & 30.90 \\[-1.0ex] 
\rule{0pt}{10pt} DABNet~\cite{dabnet} & \multicolumn{1}{|c|}{5.18} & 77.43 & \multicolumn{1}{c|}{15.27} & 81.19 & 0.00 & 0.09 & 0.00 & 4.10 & 10.49 & 0.00 & 36.18 & 42.83& 0.00 & 8.30 & 0.00  \\[-1.0ex] 
\rule{0pt}{10pt} LEDNet~\cite{lednet}  & \multicolumn{1}{|c|}{6.23} & 86.07 & \multicolumn{1}{c|}{46.40} & 88.59 & 28.13 & 36.72 & 32.45 & 43.77 & 38.55 & 41.51 & 64.19 & 60.05 & 42.40 & 53.12 & 27.29 \\
\rowcolor{gray!15}\rule{0pt}{10pt} Trans4Trans-T &  \multicolumn{1}{|c|}{10.45} &  93.23  & \multicolumn{1}{c|}{68.63} & 94.44 & 48.39 & 61.89 & 61.86 & 61.14 & 54.83 & 73.60 & 83.03 & 75.20 & 74.69 & 75.26 & 59.19  \\ 
\midrule

\rule{0pt}{10pt} ICNet~\cite{icnet}  & \multicolumn{1}{|c|}{10.64} & 78.23 &\multicolumn{1}{c|}{23.39} & 83.29 & 2.96 & 4.91 & 9.33 & 19.24 & 15.35 & 24.11 & 44.54 & 41.49 & 7.58 & 27.47 & 3.80  \\[-1.0ex] 
\rule{0pt}{10pt} BiSeNet~\cite{bisenet}  & \multicolumn{1}{|c|}{19.91} & 89.13 & \multicolumn{1}{c|}{58.40} & 90.12 & 39.54 &53.71  &50.90  & 46.95& 44.68 & 64.32 & 72.86 &63.57 &61.38 & 67.88 &44.85  \\
\rowcolor{gray!15}\rule{0pt}{10pt} Trans4Trans-S &  \multicolumn{1}{|c|}{19.92} &  94.57  & \multicolumn{1}{c|}{74.15} & 95.60 & \textbf{57.05} & 71.18 & \textbf{70.21} & 63.95 & 61.25 & 81.67 & 87.34 & 78.52 & \textbf{77.13} & 81.00 & 64.88 \\ 
\midrule

\rule{0pt}{10pt} DenseASPP~\cite{denseaspp} & \multicolumn{1}{|c|}{36.20} & 90.86 & \multicolumn{1}{c|}{63.01} & 91.39 & 42.41 & 60.93 & 64.75 & 48.97 & 51.40 & 65.72 & 75.64 & 67.93 & 67.03 & 70.26 & 49.64  \\[-1.0ex] 
\rule{0pt}{10pt} DeepLabv3+~\cite{deeplabv3+} & \multicolumn{1}{|c|}{37.98} & 92.75 & \multicolumn{1}{c|}{68.87}  & 93.82 & 51.29 &64.65  & 65.71 & 55.26 & 57.19 & 77.06 & 81.89 & 72.64 &70.81  & 77.44 & {58.63}\\[-1.0ex] 
\rule{0pt}{10pt} FCN~\cite{fcn}  & \multicolumn{1}{|c|}{42.23} & 91.65 & \multicolumn{1}{c|}{62.75} & 93.62 & 38.84 & 56.05 & 58.76 & 46.91 & 50.74 & 82.56 &78.71 & 68.78 & 57.87 & 73.66 & 46.54 \\[-1.0ex] 
\rule{0pt}{10pt} OCNet~\cite{ocnet}  & \multicolumn{1}{|c|}{43.31} & 92.03 & \multicolumn{1}{c|}{66.31} & 93.12 & 41.47 & 63.54 & 60.05 & 54.10 & 51.01 & 79.57 & 81.95 & 69.40 & 68.44 & 78.41 & 54.65\\[-1.0ex] 
\rule{0pt}{10pt} RefineNet~\cite{refinenet}  & \multicolumn{1}{|c|}{44.56} & 87.99 & \multicolumn{1}{c|}{58.18} & 90.63 & 30.62 & 53.17 & 55.95 & 42.72 &46.59 & 70.85 & 76.01 & 62.91 & 57.05 & 70.34 & 41.32   \\[-1.0ex] 
\rule{0pt}{10pt} Trans2Seg~\cite{xie2021segmenting} &  \multicolumn{1}{|c|}{49.03} &  {94.14}  & \multicolumn{1}{c|}{{72.15}} & {95.35} & 53.43 & {67.82} & 64.20 & {59.64} & {60.56} & \textbf{88.52} & {86.67} & {75.99} & {73.98} & {82.43} & 57.17 \\ [-1.0ex] 
\rule{0pt}{10pt} TransLab~\cite{xie2020segmenting}  & \multicolumn{1}{|c|}{61.31} & 92.67  & \multicolumn{1}{c|}{69.00} & 93.90 & {54.36} & 64.48 & 65.14 & 54.58 & 57.72 & 79.85 & 81.61 &72.82 & 69.63 & 77.50 & 56.43 \\ [-1.0ex] 
\rule{0pt}{10pt} DUNet~\cite{dunet}  & \multicolumn{1}{|c|}{123.69} & 90.67 & \multicolumn{1}{c|}{59.01} & 93.07 & 34.20 & 50.95 & 54.96 & 43.19 & 45.05 & 79.80 & 76.07 & 65.29 & 54.33 & 68.57 & 42.64   \\[-1.0ex] 
\rule{0pt}{10pt} U-Net~\cite{unet}  & \multicolumn{1}{|c|}{124.55} & 81.90 & \multicolumn{1}{c|}{29.23} & 86.34 & 8.76  & 15.18  & 19.02 & 27.13 & 24.73 & 17.26 & 53.40 & 47.36 & 11.97 & 37.79 & 1.77 \\[-1.0ex] 
\rule{0pt}{10pt} DANet~\cite{danet}  & \multicolumn{1}{|c|}{198.00} & 92.70  &\multicolumn{1}{c|}{68.81} & 93.69 & 47.69 & 66.05  & 70.18 & 53.01 & 56.15 & 77.73 & 82.89 & 72.24 & 72.18 & 77.87 & 56.06 \\[-1.0ex] 
\rule{0pt}{10pt} PSPNet~\cite{pspnet}  & \multicolumn{1}{|c|}{187.03} & 92.47 & \multicolumn{1}{c|}{68.23} & 93.62  & 50.33 & 64.24 & {70.19} & 51.51 & 55.27 & 79.27 & 81.93 & 71.95 &   68.91 & 77.13 & 54.43 \\

\midrule 
\rowcolor{gray!15}\rule{0pt}{10pt} Trans4Trans-M &  \multicolumn{1}{|c|}{34.38} &  \textbf{95.01}  & \multicolumn{1}{c|}{\textbf{75.14}} & \textbf{96.08} & 55.81 & \textbf{71.46} & 69.25 & \textbf{65.16} & \textbf{63.96} & 83.84 & \textbf{88.21} & \textbf{80.29} & 76.33 & \textbf{83.09} & \textbf{68.09} \\
\bottomrule
\end{tabular}
}
\caption{Computation complexity in GFLOPs and category-wise accuracy evaluation and comparison with state-of-the-art semantic segmentation methods on the Trans10K-v2 dataset~\cite{xie2021segmenting}.}
\label{tab:sota}
\end{table*}

\noindent\textbf{Comparison to state-of-the-art models.}
As shown in Table~\ref{tab:sota}, the performance of both accuracy- and efficiency-oriented semantic segmentation approaches are compared according to~\cite{xie2020segmenting}. The superiority of Trans4Trans is further confirmed through the listed experimental results in Table~\ref{tab:sota}, compared with both CNNs and transformer-based approaches like Trans2Seg~\cite{xie2021segmenting}. Our medium Trans4Trans model outperforms the state-of-the-art method Trans2Seg by $2.99\%$ in mIoU and $0.87\%$ in Acc, while requiring much less GFLOPs. For category-wise accuracy, our Trans4Trans model achieves state-of-the-art performances in IoU on the classes \emph{background}, \emph{jar or tank}, \emph{window}, \emph{door}, \emph{cup}, \emph{wall}, \emph{bottle}, and \emph{box}, indicating the efficacy of transparent object segmentation of the proposed Trans4Trans architecture.

\subsection{Segmentation Robustness in Driving Scenes}
\label{sec:driving}

\begin{table}[!t]
\centering
\resizebox{\columnwidth}{!}{
\begin{tabular}{@{}lll|cc|ll@{}}
\toprule
\textbf{Network} & \textbf{Encoder} & \textbf{Decoder} & \textbf{GFLOPs} & \textbf{MParams} & \textbf{Cityscapes} & \textbf{ACDC}\\ \midrule \midrule
PVT-T   & PVT-T~\cite{pvt} & \multirow{3}{*}{Transformer~\cite{xie2021segmenting}} &10.30  &13.11 & 58.09 & 53.65    \\
PVT-S   & PVT-S~\cite{pvt} &  &19.77  &24.35 & 59.68 & 57.13     \\
PVT-M   & PVT-M~\cite{pvt} &  &36.87  &51.83 & 60.38 & 58.60     \\ 
\midrule

Trans4Trans-T & PVT-T~\cite{pvt}    & \multirow{3}{*}{TPM / Single-head} &10.45  &12.71 & 60.41\gbf{+2.32} & 54.37\gbf{+0.72}    \\
Trans4Trans-S & PVT-S~\cite{pvt}    &  &21.98  &25.00 & 63.08\gbf{+3.40} & 60.70\gbf{+3,57}     \\
Trans4Trans-M & PVT-M~\cite{pvt}    &  &44.38  &48.77 & 65.63\gbf{+5,25} & 61.91\gbf{+3,31}     \\
\midrule

Trans4Trans-T & PVT-T~\cite{pvt}    & \multirow{3}{*}{TPM / Dual-head} &11.23  &13.10 & 57.42\rbf{-0.67} & 56.36\gbf{+2.71}    \\
Trans4Trans-S & PVT-S~\cite{pvt}    &  &24.82  &26.45 & 62.39\gbf{+2.71} & 62.14\gbf{+5.01}     \\
Trans4Trans-M & PVT-M~\cite{pvt}    &  &55.16  &54.28 & 63.00\gbf{+2.62} & 63.88\gbf{+5.28}     \\
\midrule[1.1px]
Trans4Trans-T & PVTv2-B1~\cite{wang2021pvtv2}    & \multirow{3}{*}{TPM / Single-head} & 9.18 & 13.53 & 63.25\gbf{+5.16} & 59.25\gbf{+5.60}   \\
Trans4Trans-S & PVTv2-B2~\cite{wang2021pvtv2}    &  & 19.27 & 25.62 & 67.28\gbf{+7.60} & 64.61\gbf{+7.48}    \\
Trans4Trans-M & PVTv2-B3~\cite{wang2021pvtv2}    &  & 41.89 & 49.55 & 69.34\gbf{+8.96} & 65.92\gbf{+7.32}    \\
\midrule
Trans4Trans-T & PVTv2-B1~\cite{wang2021pvtv2}    & \multirow{3}{*}{TPM / Dual-head} & 10.00 & 13.93 & 62.31\gbf{+4.22} & 61.86\gbf{+8.21}    \\
Trans4Trans-S & PVTv2-B2~\cite{wang2021pvtv2}    &  & 22.17 & 27.08 & 65.98\gbf{+6.30} & 64.83\gbf{+7.70}     \\
Trans4Trans-M & PVTv2-B3~\cite{wang2021pvtv2}    &  & 52.77 & 55.09 & 69.05\gbf{+8.67} & 66.65\gbf{+8.05}   \\
\bottomrule[1px]
\end{tabular}
}
\caption{\small Effectiveness of our model on driving scene datasets, \eg Cityscapes~\cite{cityscapes} and ACDC~\cite{sakaridis2021acdc}. The first group is selected as the baseline. The fourth and fifth groups further verify the flexibility of our TPM adapted to other backbones. 
\#MParams, \#GFLOPs are calculated with the input size of $512{\times}512$. The embedding dimensions of \emph{model}-T, -S, and -M decoders are $\{64, 128, 256\}$.
}
\label{tab:performance_driving}
\end{table}

Apart from segmenting general and transparent objects, we further verify our Trans4Trans model on driving scene datasets to show its potential for various ITS applications.

\noindent\textbf{Ablation of dual-head Trans4Trans.}
Five groups of results are shown in Table~\ref{tab:performance_driving}. Our TPM-based Trans4Trans trained at $512{\times}512$ resolution, illustrates better performance compared with PVT on Cityscapes and ACDC datasets focusing on driving scenes. On Cityscapes, Trans4Trans-M leveraging PVT encoder outperforms PVT-M by $5.25\%$ and Trans4Trans-M leveraging PVTv2 as the encoder surpasses by $8.96\%$ (with an additional $3.71\%$ gain). They both utilize TPM/Single-head as the decoder in Trans4Trans, indicating the learning capacity of our proposed approach for driving scene understanding. On ACDC, our Trans4Trans-M leveraging PVT encoder outperforms PVT-M by $5.28\%$ and the one leveraging PVTv2-M as the encoder exceeds by $8.05\%$ (with an additional $2.77\%$ boost) while utilizing TPM/Dual-head in the decoder architecture. Since ACDC is a dataset containing different adverse conditions, these results evidence that TPM/Dual-head has the better robustness under environment changes in driving scene segmentation, as it incorporates more generalized knowledge learned from diverse images in both datasets.

\begin{table}[!t]
\centering
\resizebox{.99\columnwidth}{!}{
\begin{tabular}{c|c|rrc}
    \toprule
    {\textbf{Methods}}&{\textbf{Encoder}}&{\textbf{GFLOPs~$\downarrow$}}&{\textbf{MParams~$\downarrow$}}&\textbf{mIoU~(MS)~$\uparrow$}\\
    \midrule	
    Fast-SCNN~\cite{fastscnn}  & Fast-SCNN & 2.07 & 1.46 & 72.65\\    
    CGNet~\cite{wu2021cgnet}  & CGNet-M3N21 & 7.72 & 0.50 & 64.80\\
    \rowcolor{gray!15} Trans4Trans-T    & PVTv2-B1~\cite{wang2021pvtv2}& 20.66 & 13.53 & 78.23 \\
    SegFormer-B1~\cite{segformer}    & MiT-B1 &  29.85 & 13.66 & 78.43 \\
    ERFNet~\cite{romera2017erfnet} &ERFNet & 30.22 & 2.07 & 72.10 \\
    \midrule
    \rowcolor{gray!15} Trans4Trans-S    & PVTv2-B2~\cite{wang2021pvtv2}& 43.37 & 25.62 & 80.02 \\
    PSPNet~\cite{pspnet}          & MobileNetV2 &  119.09 & 13.72 & 70.20 \\	
    PSPNet~\cite{pspnet}          & ResNet-18   &  119.27 & 12.77 & 76.90 \\
    SegFormer-B2~\cite{segformer}    & MiT-B2      & 127.86 & 27.33 & 80.46 \\
    SegFormer-B3~\cite{segformer}    & MiT-B3      & 160.78 & 47.18 & 81.50 \\
    DeepLabv3+~\cite{deeplabv3+}       & MobileNetv2 & 169.53 & 18.70 & 75.20 \\
    EMANet~\cite{li2019expectation}          & ResNet-50   & 379.00 & 42.09 & 80.49 \\
    PSPNet~\cite{pspnet}          & ResNet-50   &  401.51 & 48.98 & 79.96 \\
    DNL~\cite{yin2020disentangled}             & ResNet-50   & 449.73 & 50.02 & 80.70 \\
    EMANet~\cite{li2019expectation}          & ResNet-101  & 553.79 & 61.08 & 81.00 \\
    PSPNet~\cite{yin2020disentangled}          & ResNet-101  &  573.48 & 67.95 & 80.04 \\
    DNL~\cite{yin2020disentangled}             & ResNet-101  & 624.52 & 69.02 & 80.68 \\
    SETR-Naive~\cite{setr} & ViT-L~\cite{vit} & 698.52 & 306.58 & 77.90\\
    SETR-MLA~\cite{setr} & ViT-L~\cite{vit} & 712.76 & 310.81 & 77.24 \\
    SETR-PUP~\cite{setr} & ViT-L~\cite{vit} & 818.26 & 319.11 & 79.34 \\
    \midrule
    \rowcolor{gray!15} Trans4Trans-M    & PVTv2-B3~\cite{wang2021pvtv2}& 94.25 & 49.55 & \textbf{81.54} \\
    \bottomrule
\end{tabular}}
\caption{\small Comparison of state-of-the-art models on Cityscapes~\cite{cityscapes}. Methods in the first group are designed for efficient semantic segmentation and have smaller \text{\#GFLOPs}. \#MParams, \#GFLOPs are calculated with the input size of $768{\times}768$. ``MS'' means multi-scale testing.}
\label{tab:sota_cs}
\end{table}

\noindent\textbf{Segmentation in normal conditions.}
As shown in Table~\ref{tab:sota_cs}, experimental results of our proposed Trans4Trans approach trained with the input size of $768{\times}768$, together with state-of-the-art semantic segmentation methods are presented\footnote[1]{For a fair comparison, model weights are obtained by the same framework MMSegmentation: https://github.com/open-mmlab/mmsegmentation}. Here, all the decoders for Trans4Trans are constructed as the single-head decoder for the experiments on Cityscapes according to the best performance of single-head decoder illustrated in Table~\ref{tab:performance_driving} on Cityscapes. Our Trans4Trans-M approach with PVTv2-B3 as encoder outperforms the others and achieves the best performance with an mIoU of $81.54\%$ on Cityscapes, whose images are collected under normal weather and favorable illumination conditions. Compared with the state-of-the-art methods such as SETR~\cite{setr} and PSPNet~\cite{pspnet}, our Trans4Trans approach shows smaller GFLOPs ($94.25$) and less parameters ($49.55M$), which are relevant for fast inference in automated vehicles. Our TransTrans models with lighter encoder architectures indicated as Trans4Trans-T and Trans4Trans-S also show high scores of $78.23\%$ and $80.02\%$ in mIoU when leveraging PVTv2-B1 and PVTv2-B2 as the encoder. The lightest Trans4Trans outperforms state-of-the-art efficient networks FastSCNN~\cite{fastscnn} and ERFNet~\cite{romera2017erfnet} by large margins, and it achieves a similar score as SegFormer~\cite{segformer} while being significantly more efficient.

\begin{table}[!t]
\centering
\resizebox{.99\columnwidth}{!}{
\begin{tabular}{lcr|cccc|cc}
    \toprule
    \textbf{Method} & \textbf{Trained on} & \textbf{GFLOPs}~$\downarrow$ & \textbf{Fog} & \textbf{Night} & \textbf{Rain} & \textbf{Snow} & \textbf{All-ACDC} & \textbf{All-DADA} \\
    \midrule
    DeepLabv3+~\cite{deeplabv3+}&{CS} & 178.1 & 45.7 & 25.0 & 50.0 & 42.0 & 41.6 & 10.4\\
    HRNet~\cite{hrnet} &CS & 210.5 & 38.4 & 20.6 & 44.8 & 35.1 & 35.3 & 15.5\\
    \rowcolor{gray!15} Trans4Trans-M &CS & 41.8 & \textbf{74.1} & \textbf{31.1} & \textbf{63.4} & \textbf{57.9} & \textbf{55.7} & \textbf{27.7} \\ 
    \midrule    
    DeepLabv3+~\cite{deeplabv3+} & ACDC & 178.1 & 69.1 & 60.9 & 74.1 & 69.6 & 70.5 & 26.8\\
    HRNet~\cite{hrnet} & ACDC & 210.5 & 74.7 & \textbf{65.3} & \textbf{77.7} & 76.3 & 75.0 & 27.5 \\
    \rowcolor{gray!15} Trans4Trans-M & ACDC & 41.8 & \textbf{79.8} & 55.3 & 77.4 & \textbf{78.6} & \textbf{75.2} & \textbf{32.4}\\
    \midrule    
    \rowcolor{gray!15} Trans4Trans-M & ACDC+CS & 41.8 & \textbf{81.4} & 56.0 & 77.0 & \textbf{78.8} & \textbf{76.3} & \textbf{39.2}  \\
    \bottomrule
\end{tabular}}
\caption{\small Comparison of different semantic segmentation models on adverse~(Fog, Night, Rain, Snow, and All-ACDC~\cite{sakaridis2021acdc}) and extreme accident~(All-DADA~\cite{zhang2021issafe}) conditions.
CS: Cityscapes~\cite{cityscapes}.
\#GFLOPs are calculated with the input size of $768{\times}768$.}
\label{tab:driving_robust}
\end{table}

\begin{figure*}[!t]
    \centering
    \includegraphics[width=0.99\linewidth]{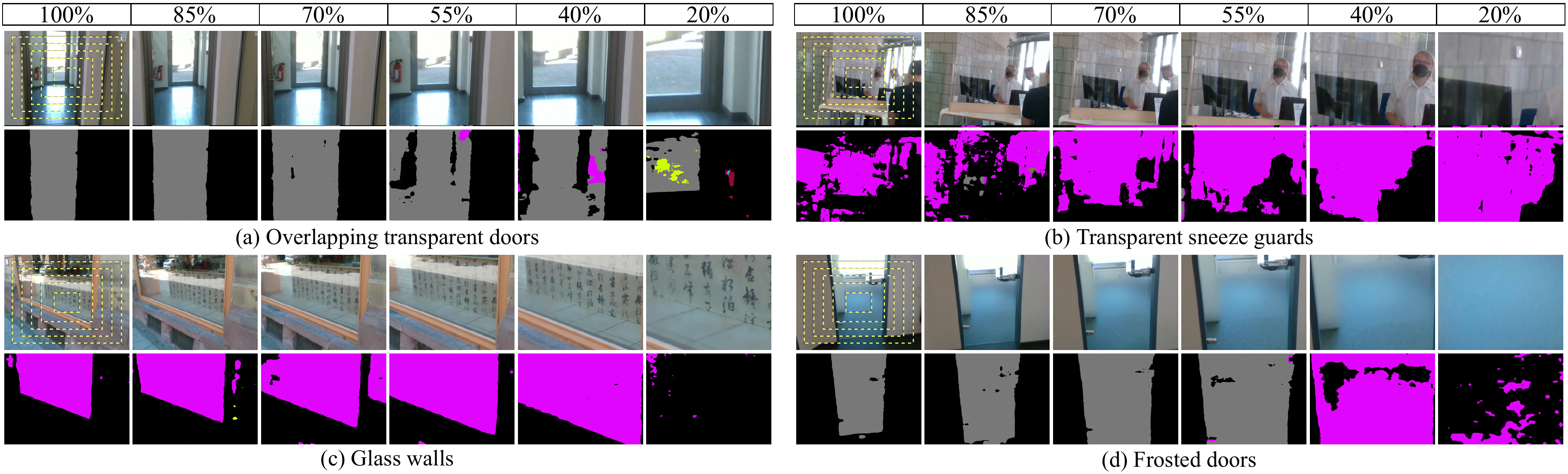}
    \caption{\small Visualization of segmentation results from different cropped regions based on image center. Images of six scales from 100\% to 20\% are cropped from its original images and are separately segmented, in order to ablate the effect of image context.
    }
    \label{fig:vis_cropped}
\end{figure*}

\noindent\textbf{Segmentation in adverse conditions.}
As shown in Table~\ref{tab:driving_robust}, we further test the performance our proposed Trans4Trans-M approach on ACDC~\cite{sakaridis2021acdc} and DADA-seg~\cite{zhang2021issafe} in adverse conditions and extreme accident scenes individually. The results of Trans4Trans are obtained via MMSegmentation with a resolution of $768\times768$. The first three rows of models are trained on Cityscapes in normal and favorable conditions and tested on ACDC and DADA-seg datasets. Our Trans4Trans-M indicates higher performances of $55.7\%$ and $27.7\%$ in mIoU when compared with HRNet~\cite{hrnet} and DeepLabV3+~\cite{deeplabv3+}. Trans4Trans performs clearly better than them in various adverse conditions including \emph{foggy, nighttime, rainy, snowy}, and extreme \emph{accident} scenes, which demonstrates its high generalization capacity to these unseen domains. This is because with both transformer-based encoder and decoder, Trans4Trans can associate long-range visual concepts for robustly inferring semantics, despite local texture- and illumination changes in different scenarios like nighttime and accident scenes. Then, the following three rows of models are trained on ACDC and tested on ACDC and DADA-seg. Our Trans4Trans again indicates better overall performances with $75.2\%$ and $32.4\%$ on All-ACDC and -DADA. Finally, we train Trans4Trans by merging ACDC and Cityscapes which have the same $19$ classes, and the learned model shows the best overall scores on All-ACDC and All-DADA with $76.3\%$ and $39.2\%$ in mIoU, respectively, illustrating that co-training on normal and adverse data can improve the performance of the model under both adverse and extreme accident conditions.

\subsection{Real-time Performance}
\label{sec:realtime}
To measure the inference speed of the different versions of our dual-head Trans4Trans model, $300$ samples from the Trans10K-v2 test set with a batch size of $1$ and a resolution of $512{\times}512$ are tested on three different GPUs, \ie, a mobile NVIDIA AGX Xavier in the MAXN mode, an NVIDIA GeForce MX350 on a lightweight laptop, and an RTX 2070 on a workstation. As shown in Table~\ref{tab:inf_time}, the running time (latency) of our tiny Trans4Trans model on three GPUs are considerably lower than the other two versions, meanwhile the performances of the three models on both datasets are suitable for our system. In real applications, the more timely response of the navigation system is beneficial for assisting users with a similar prediction accuracy on each frame. Hence, the tiny version model is selected. The model is deployed on the lightweight laptop (with an MX350 GPU) to conduct the user study. Regarding the entire system, it takes $0.14 ms$ ($\pm 0.11$) for the image acquisition step, $101.5 ms$ ($\pm 0.3$) for semantic segmentation, and $74.47 ms$ ($\pm 2.41$) for the assistive algorithm. Thus, the entire system requires a total of $176 ms$. 

\begin{table}[t]
\centering
\resizebox{\columnwidth}{!}{
\begin{tabular}{l|ccc}
    \toprule
    {\textbf{Network}}&\textbf{NVIDIA Xavier~(ms)~$\downarrow$}&{\textbf{MX350~(ms)~$\downarrow$}}&{\textbf{RTX 2070~(ms)~$\downarrow$}}\\
    \midrule
    Trans4Trans-M & 115.9(\textpm 1.1) / 202.8(\textpm 1.1)& 186.1(\textpm 0.3) / 243.2(\textpm 0.3) & 22.9(\textpm 0.3) / 36.6(\textpm 0.8) \\
    Trans4Trans-S & 95.3(\textpm 0.6) / 158.6(\textpm 1.8)& 140.6(\textpm 0.3) / 188.4(\textpm 0.4) & 17.1(\textpm 0.3) / 27.7(\textpm 0.5) \\
    Trans4Trans-T & 75.8(\textpm 0.7) / 122.7(\textpm 0.7) & 101.5(\textpm 0.3) / 141.7(\textpm 1.6) & 12.8(\textpm 0.5) / 20.3(\textpm 0.5)  \\
    
    \bottomrule
\end{tabular}}
\caption{\small Inference time~(ms/frame) of dual-head Trans4Trans is tested in half-/single-precision on various GPUs at $512{\times}512$.}
\label{tab:inf_time}
\end{table}

\subsection{Transparent Feature}
\label{sec:feature}
\noindent\textbf{Analysis of contexts and reflections.}
In order to ablate the impact of context information (such as door frames and walls) or reflection cues on the segmentation of transparent objects, we visualize the real-world segmentation results from different scales (six scales from $100\%$ to $20\%$) of the original images in Fig.~\ref{fig:vis_cropped}. The segmentation results are all generated by Trans4Trans. In Fig.~\ref{fig:vis_cropped}(a), among all six scales, even in the $20\%$ case with less context, the segmentation of two overlapping doors are accurately obtained. In Fig.~\ref{fig:vis_cropped}(b), the challenging sneeze guards with less frame cues are accurately segmented, since this sneeze guard has reflection characteristics as well. In the case of $40\%$ ratio in Fig.~\ref{fig:vis_cropped}(c), with only a one-side outer frame and part of the reflection, it can still segment the area of the glass wall. However, in the $20\%$ ratio, it is confused due to the tiny frame and absence of any reflections. The incomplete segmentation and errors in the latter two ratios of Fig.~\ref{fig:vis_cropped}(d) are caused by the lack of texture information and reflections. Based on the analysis of these visualization results, two insights are provided: (1) The contextual information, \eg the outer frame, is a vital factor for the segmentation of transparent objects; (2) The reflection characteristic of glass or transparent objects is also crucial. Benefiting from the symmetrical encoder-decoder architecture, our Trans4Trans model can robustly segment transparent objects even with diminishing context cues in most of the complex real-life scenes.

\begin{figure*}[!t]
    \centering
    \includegraphics[width=0.99\linewidth]{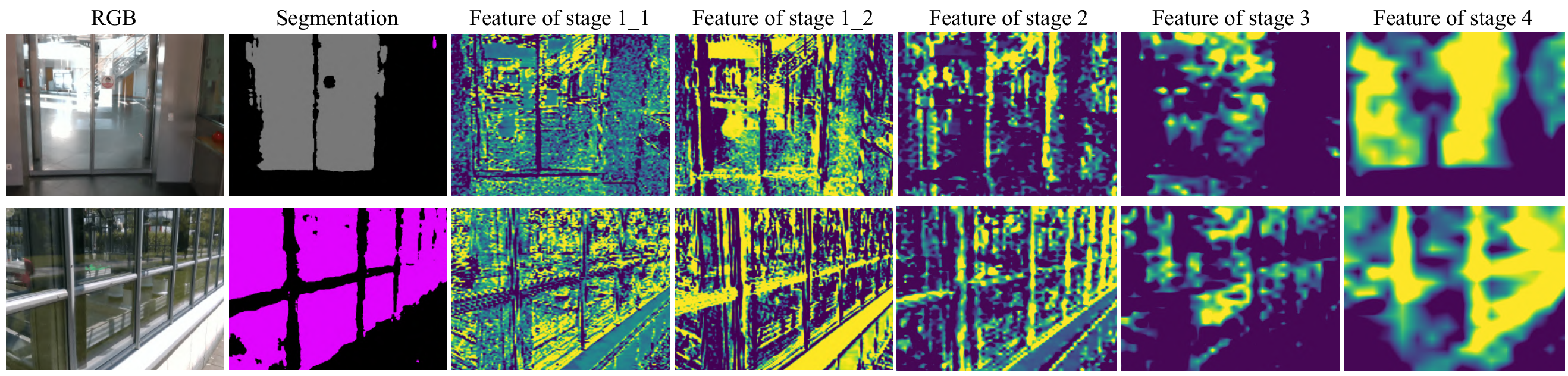}
    \caption{\small Visualization of feature maps, corresponding to four different stages in the TPM decoder. The two feature maps from the same stage (stage 1\_1 and stage 1\_2) indicate the global and local activated features of transparent objects.}
    \label{fig:vis_feat_map}
\end{figure*}

\noindent \textbf{Features parsing.}
For semantic segmentation, the local feature from the low-level layer is often preserved for the fine prediction such as an exact boundary of the detected object, while the contextual feature from the high-level layer is for distinguishing the object category. 
To investigate the feature parsing capability of the TPM decoder, we visualize the interpreted feature maps from four stages of the decoder, as shown in Fig.~\ref{fig:vis_feat_map} with real-world scenes. The results are generated by the single-head Trans4Trans trained on Trans10K-v2. We get two observations: (1) In various low- and high-level stages, the parsed features contain both fine-grained and contextual information, thanks to our Trans4Trans architectural design which enables to capture long-range context priors from very first layers. For example, the fine feature in the stage 1\_2 of the first row contains the instance-level information such as \emph{glass door}. The contextual feature in stage 4 contains the precise boundary of the \emph{glass door} as well. Both types of features are critical for the segmentation task; (2) In the same stage, the fine-grained feature is simultaneously reflected in the object and its surrounding. For example, the feature in stage 1\_1 of the first row is more inclined to the surrounding, and the one in stage 1\_2 is to the area of the \emph{glass door}, and vice versa in the second row.

\subsection{Qualitative Segmentation Analysis}
\label{sec:qualitative}

\begin{figure*}[!ht]
    \centering
    \includegraphics[width=0.99\linewidth]{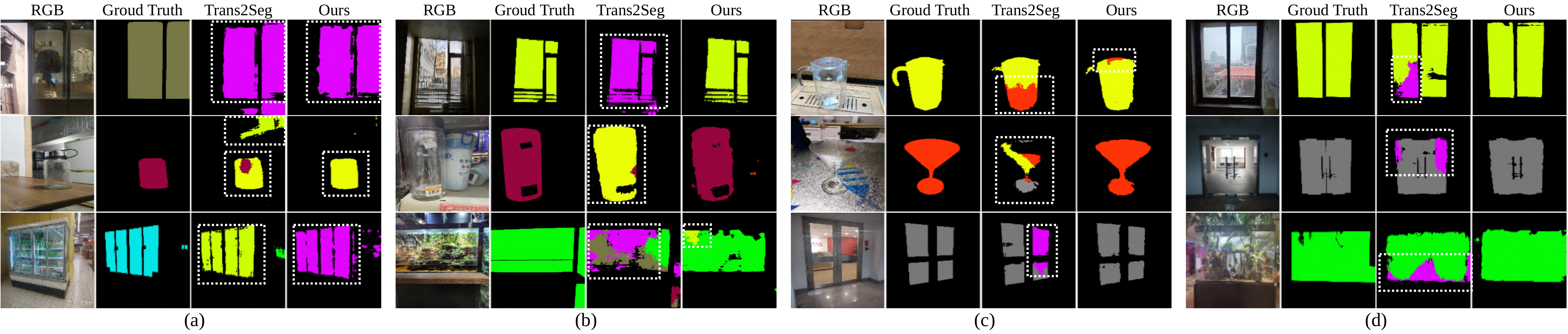}
    \caption{\small Qualitative analysis on Trans10K-v2 test set. (a) shows some negative predictions from both models. In (b), our Trans4Trans can correctly segment those cases failed by Trans2Seg. In (c) and (d), our results are more precise.
    }
    \label{fig:vis_testset}
\end{figure*}

\noindent\textbf{Visualization of transparency segmentation.}
Fig.~\ref{fig:vis_testset} shows qualitative comparisons between our Trans4Trans-Tiny and the previous state-of-the-art approach Trans2Seg~\cite{xie2021segmenting}. Fig.~\ref{fig:vis_testset}(a) illustrates some failed recognition cases of both models, but our model can yield a clearly better boundary distinction. Fig.~\ref{fig:vis_testset}(b) shows examples where our model predicts the correct label, but Trans2Seg is confused, indicating the reliable performance of our proposed approach. In Fig.~\ref{fig:vis_testset}(c)(d), it can be seen that our model is not only effective in detecting navigation-critical \emph{glass door} and \emph{glass window}, but can also predict more refined segmentation of small objects like \emph{jar/tank} and \emph{glass cup}.

\begin{figure}[!t]
    \centering
    \includegraphics[width=0.99\linewidth]{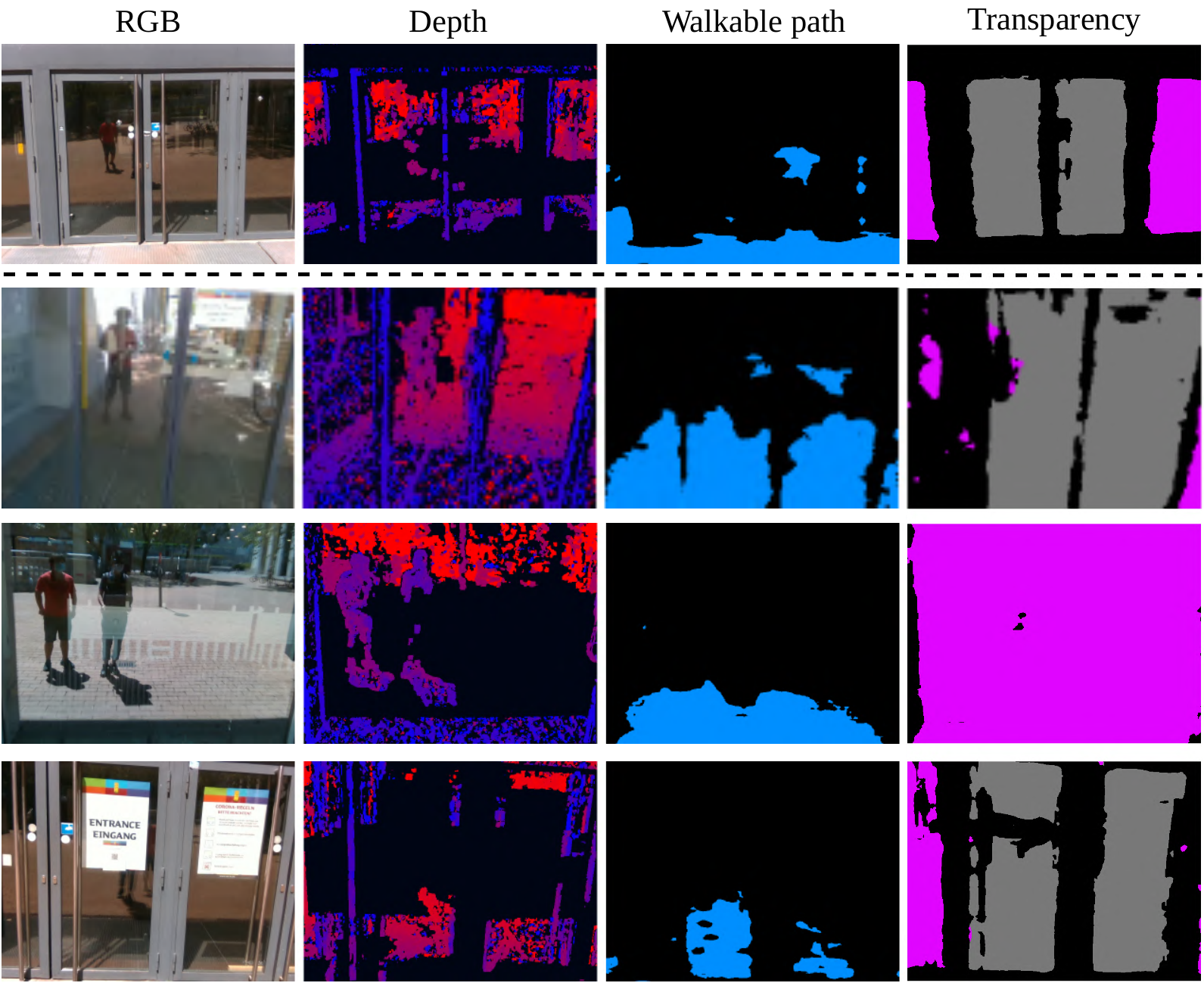}
    \caption{\small Visualization of real-world scenes. From left to right are RGB and depth image, segmentation as \textcolor{blue}{\emph{\textbf{walkable path}}} by single-head model trained on Stanford2D3D, and as transparent objects (\textcolor{gray}{\emph{\textbf{glass door}}} or \textcolor{mask_red}{\emph{\textbf{glass wall}}}) corrected by our dual-head Trans4Trans model.}
    \label{fig:vis_realworld}
\end{figure}

We further perform field tests by navigating around the university campus and capturing real-world scenes with our smart vision glasses. The collected RGB-D images and corresponding predictions are shown in Fig.~\ref{fig:vis_realworld}. Meanwhile, the semantic segmentation runs on the mobile GPU processor ported with our efficient vision transformer model. We visualize multiple sets of predictions as shown in Fig.~\ref{fig:vis_realworld}(c)(d). The glass door in the first row captured at a moderate distance can be correctly distinguished from navigable areas, whereas in the other rows they are mis-classified as walkable paths by the general object segmentation model. As it can be seen, transparent surfaces are often texture-less and the infrared patterns projected by the glasses will transmit the glass regions, and thereby the depth information are often sparse, noisy, or even lost, making transparent objects a constant threat for 3D vision-based systems~\cite{aladren2014navigation,wang2017enabling} designed for helping avoid obstacles. Only learning from a common segmentation dataset can help identify the walkable areas with predicted traversable classes, but this will miss the detection of transparent obstacles, and thereby renders conventional image segmentation-based systems~\cite{yang2018unifying,lin2019deep} less effective in these scenarios, which are omnipresent in real-world navigation. In contrast, our Trans4Trans accurately and completely segments those transparent objects, meanwhile covers general objects and walkable paths, and thereby it is ideally suitable for assistance systems.

\begin{figure*}[!t]
    \centering
    \includegraphics[width=0.99\linewidth]{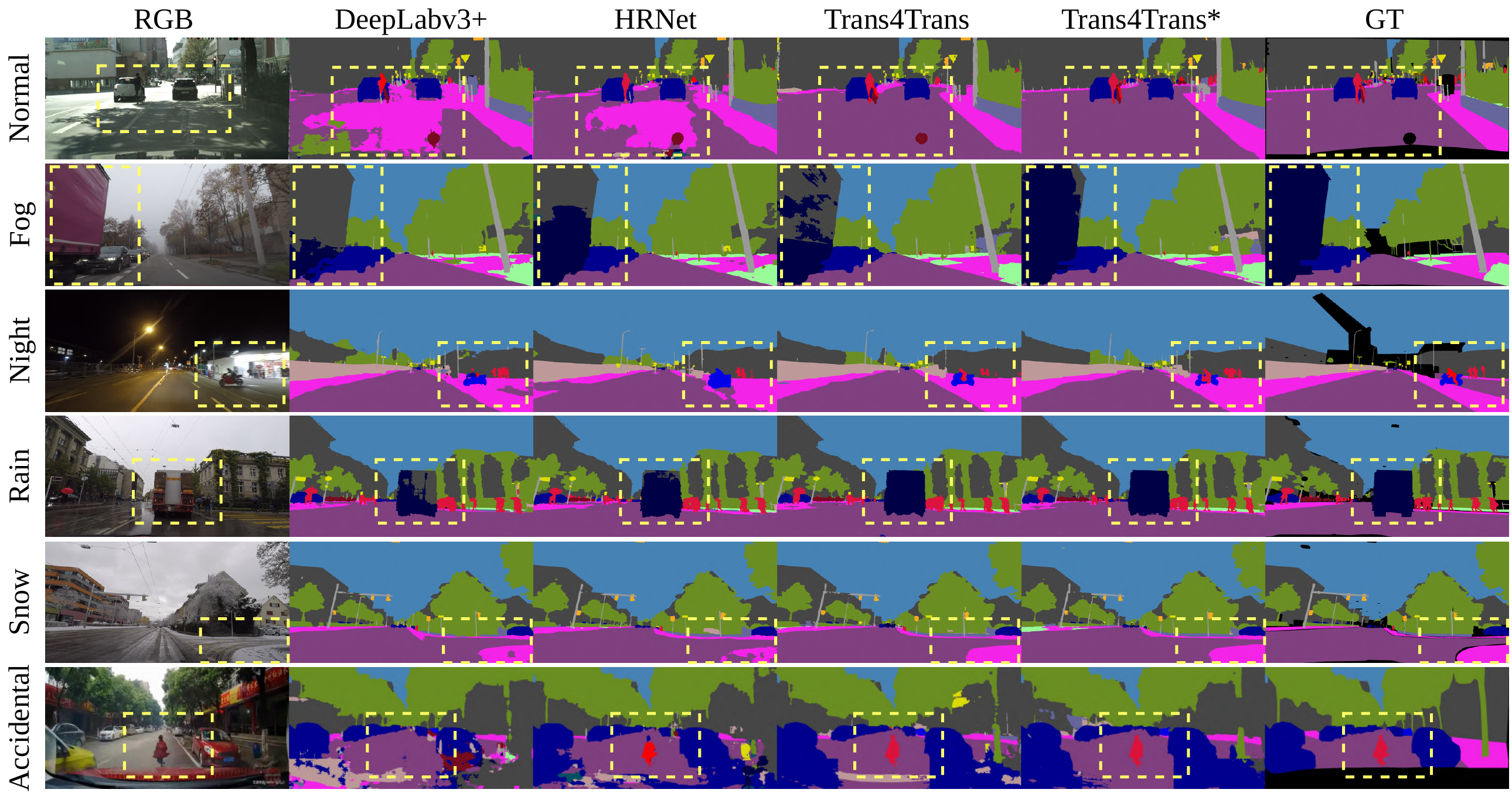}
    \caption{\small Qualitative analysis on Cityscapes~\cite{cityscapes} (\emph{Normal}), ACDC~\cite{sakaridis2021acdc} (\emph{Fog, Night, Rain}, and \emph{Snow}), and DADA-seg~\cite{zhang2021issafe} (\emph{Accidental}). The Trans4Trans* is trained on ACDC+Cityscapes, whereas other models are trained on ACDC dataset. 
    }
    \label{fig:vis_driving}
\end{figure*}

\noindent\textbf{Visualization of driving scene segmentation.}
As shown in Fig.~\ref{fig:vis_driving}, we visualize the predictions of Trans4Trans* trained on Cityscapes+ACDC, in comparison to DeepLabv3+~\cite{deeplabv3+}, HRNet~\cite{hrnet}, and our Trans4Trans models only trained on ACDC. It can be seen that DeepLabv3+ and HRNet often produce noisy segmentation results in complex real-world conditions, like the \emph{cars} in challenging shadow and illumination situations (the first row). In adverse weather and illumination conditions, the previous methods also yield less precise and even fragmented semantics, like the \emph{trucks} in foggy and rainy scenes (the second and fourth rows) and the \emph{sidewalks} in night and snowy scenes (the third and fifth rows). In extreme accident scenes, which are safety-critical for automated vehicles, existing state-of-the-art models cannot generate reliable predictions to be propagated to upper-level applications, as the close \emph{pedestrian} is even completely recognized as \emph{road}. In contrast, Trans4Trans, which learns to gather long-range dependency from the very first layers, delivers more robust segmentation in various scenes, as it is less affected by local texture and illumination changes. Trans4Trans trained on both adverse and normal datasets further improves the performance, resulting sharp and fine-grained semantic segmentation, which is suitable for self-driving applications.

\section{User Study}
We conducted a qualitative study in order to ``understand people's needs and the context within which [our] future technology might be used'' \cite{Blandford2016_qualitative}. Another goal was to draw design conclusions for future works.
Because system-level performance and efficiency have been evaluated in Sec.~\ref{sec:realtime}, we did not evaluate any additional objective metrics, but focus on user comments and suggestions on the system.
Since the target user group is a very heterogeneous one, and participants with visual impairments are difficult to recruit, we decided to evaluate with experts in accessibility and assistive systems with focus on visual impairment.

\noindent\textbf{Methodology.}
The hardware consisted of the smart vision glasses and
a backpack with a lightweight laptop and a battery pack inside. The system's battery life under these conditions was approximately $4$ hours.

As mentioned in Algorithm~\ref{algo:system} and Sec.~\ref{sec:sys_algo},
three main functions are partitioned into four steps as mutually exclusive: (1) obstacle avoidance based on depth information (\ie $<1.0 m$) has the highest priority. (2) Three different types of transparent stuff (\emph{wall}, \emph{door}, and \emph{window}) will be alerted via speeches. (3) Walkable path will be indicated in three different directions (\emph{left}, \emph{forward}, and \emph{right}). (4) Other general objects and transparent things, listed in Sec.~\ref{sec:settings}, will be fed back.

\begin{figure}[b]
    \centering
    \includegraphics[width=0.99\linewidth]{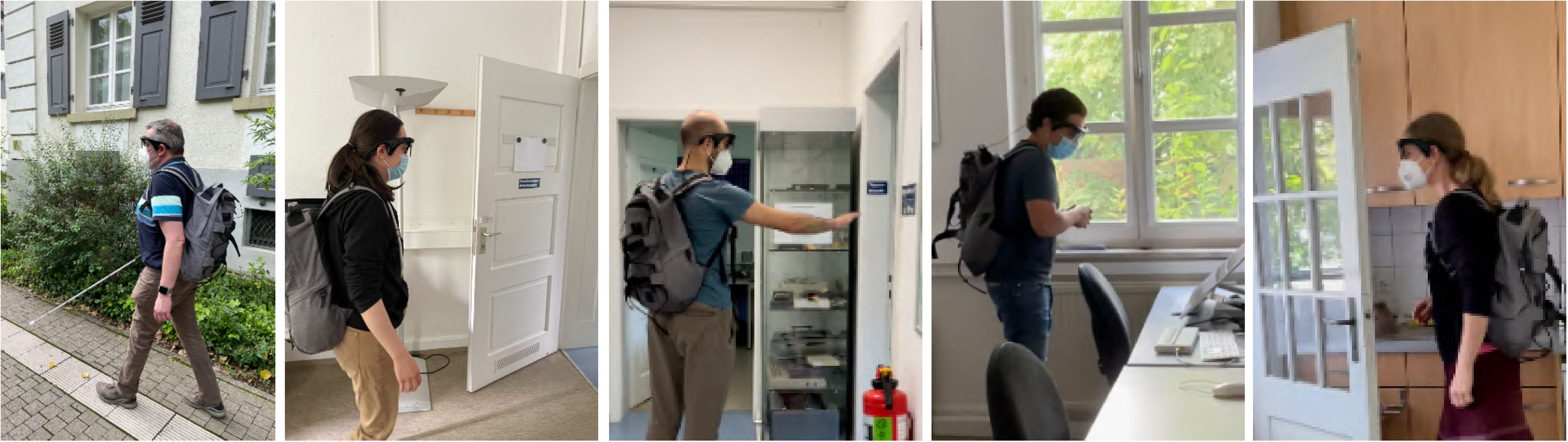}
    \caption{\small Incidences of participants using the system for navigation outdoors and indoors.}
    \label{fig:participants}
\end{figure}

Participants tried the system inside $2$ buildings (one building is mostly glazed), and the blind expert (E1B) also walked with the system on a $700 m$ route outdoors - see Fig.~\ref{fig:participants}.
The study lasted between 30-45 minutes (E5-E8) and 90 minutes (E1B).
As Corona-protective measures, everyone wore FFP2 or surgery masks throughout the entire
study 
and the prototype was disinfected several times.

After a short introduction, the participants put on the system and walked around the rooms, thinking out loud \cite{johnstone2006using}.
At the end, demographics and NASA Raw Task Load Index (RTLX)~\cite{hart2006nasa} questionnaires were filled in.

\noindent\textbf{Participants.} 
In a first step, we evaluated with 5 participants \cite{zhang2021trans4trans}, one of whom was an expert, and another one was expert and blind user at the same time. We subsequently repeated the experiment with 3 further sighted experts, and we only report here the aggregated results from the 5 experts: E1B (early blind expert), E5-E8 (sighted experts).
When asked if they can see glass objects,
E1B said he can sometimes see some light-dark contrasts, which allows him to perceive
closed windows.
Windows that open inside the room, however, are very dangerous, according to E1B, as one can get serious head injuries. All sighted participants said they can see glass objects, but some of them, like glass doors, glass walls or windows, can be challenging under particular conditions (E5).

\noindent\textbf{Cognitive load.}
The RTLX, averaged over the five expert participants, was $16.3$ with a standard deviation of $8.1$. The range is from $0$ to $100$, the lower the better. This score is enough to keep the user motivated, while not burdening too much \cite{martinez2020helping}. 
This score, however, must be critically interpreted, since it might not be representative for the users wearing the system in their daily activities. Instead, this score might reflect the cognitive load of the experts assessing the system,
since this was their task, and not simulating user behavior. Only the score of the blind participant is highly relevant for the cognitive load of users wearing the system. This score is $13.3$, thus very close to the average, but being alone, it has hardly any statistical relevance. More studies will have to be performed in the future to assess the cognitive load of the users wearing the system.

\noindent\textbf{User comments.}
A thematic analysis~\cite{Braun2006using} performed on the comments made by the experts (both recorded and from the questionnaires) yielded the following insights:

\emph{Functionality}. All experts found the system useful and were impressed by its functionality, for instance,
\emph{``for the first time, I had the feeling that artificial intelligence can be useful [...]. [I liked] how much it recognized correctly. [...] Systems react much better [than 10 years ago...]. I think it's just cool!''} (E1B);
Most positive comments are on the amount and type of objects recognized (E1B, E5, E7, E8).
The experts gave some important suggestions on subsequent system development, such as identifying more objects (E1B, E5), mounting a second camera to detect low-lying obstacles (E6), and hinting the directions of detected objects (E1B, E5-E7).
Two experts (E5, E7) commented positively upon the free path detection, and mentioned that the obstacle detection should be improved. 
Most issues with the obstacle detection came from the $2 seconds$ cycles (frame aggregation), which often caused a delay and delay inconsistencies (E1B, E5, E6, E7). To tackle this problem, it is desirable to further decrease the system response time. Regarding the suggestions from E5 and E8, adaptive feedback cycles for different functions can be implemented.
For example, the feedback of obstacle detection should be given generally faster than for the other two functions. The default in this case could be for instance $1 second$ instead of $2$.
Besides, the distances of detected objects are helpful for keeping social distances in COVID-19 pandemic times (E6, E8).
Three experts (E5-E7) considered that this system is a nice complement to the white cane, but should not be used as alternative.
    
\emph{Hardware.} 
The hardware was perceived as quite light weight (E5, E6), and in any case much better than previous prototypes (E1B) tried out by the experts in the past (at least three out of five had tried similar prototypes in the past). However, two experts (E1B, E5) considered the hardware still too big for a real-world deployment: \emph{``[use] a belt instead of a backpack [and] Bluetooth instead of cables for the glasses''}  (E1B), \emph{``ideally, it should run on a phone''} (E5).
The camera was perceived by E1B as very comfortable to wear, \emph{``although it is thick''}.
Experts E5 and E6 commented positively on the system's battery life.

\emph{Interface}. Four out of five experts thought the interface was very intuitive. Only E8 was neutral with respect to this. E6 liked \emph{``the object announcement. The acoustic signal is easy to follow + easy to understand''}. E1B thought that \emph{``the synthetic voice was very helpful, because it differentiates well from background noise''}. E8 thought that the speech output was appropriate for objects recognition, but for the other two functions, some alternatives would be better, in order to diminish the user's cognitive load.
He suggested using vibrotactile feedback for obstacle avoidance and sonification for walkable path detection. One could modulate volume or intensity and use different patterns, while not changing the frequencies - for both vibration and sonification.

\emph{Context of use}. Expert E7 thought the system is good for getting an overview of a new room, but not so good for known rooms. He also suggested implementing a couple of new functions, one to search for objects,
and one for counting objects.
Both E5 and E7 thought the system can be used for social distancing, but referred to two different functions of the system, namely obstacle detection and free path recognition.
E5 suggested to use the system also for sighted people for warning when walking while looking at the phone. 

\emph{Control}. E7 thinks the user should be in full control of the system, like it is the case with the white cane: \emph{``I can't interact with the system (mute) - the white cane does what I want''}. Both E5 and E8 thought it is important to have the option to turn functions on and off, or switch to different modes (E8).

\emph{Conclusion}. The functionality offered by the system so far can be of great use to people with visual impairments. The experts were positive about the system. Especially the object recognition was appreciated. Some improvements (such as conveying to the user the distance and direction of objects, reducing the delay for obstacle avoidance) were suggested. Also important, the users should be able to configure the system as much as possible and turn functions on and off as they need, or change the way things are conveyed (speech, sonification, vibration). Due to the heterogeneity of the user group, the configurability is a very important aspect. 

\begin{figure}[h]
    \centering
    \includegraphics[width=0.99\linewidth]{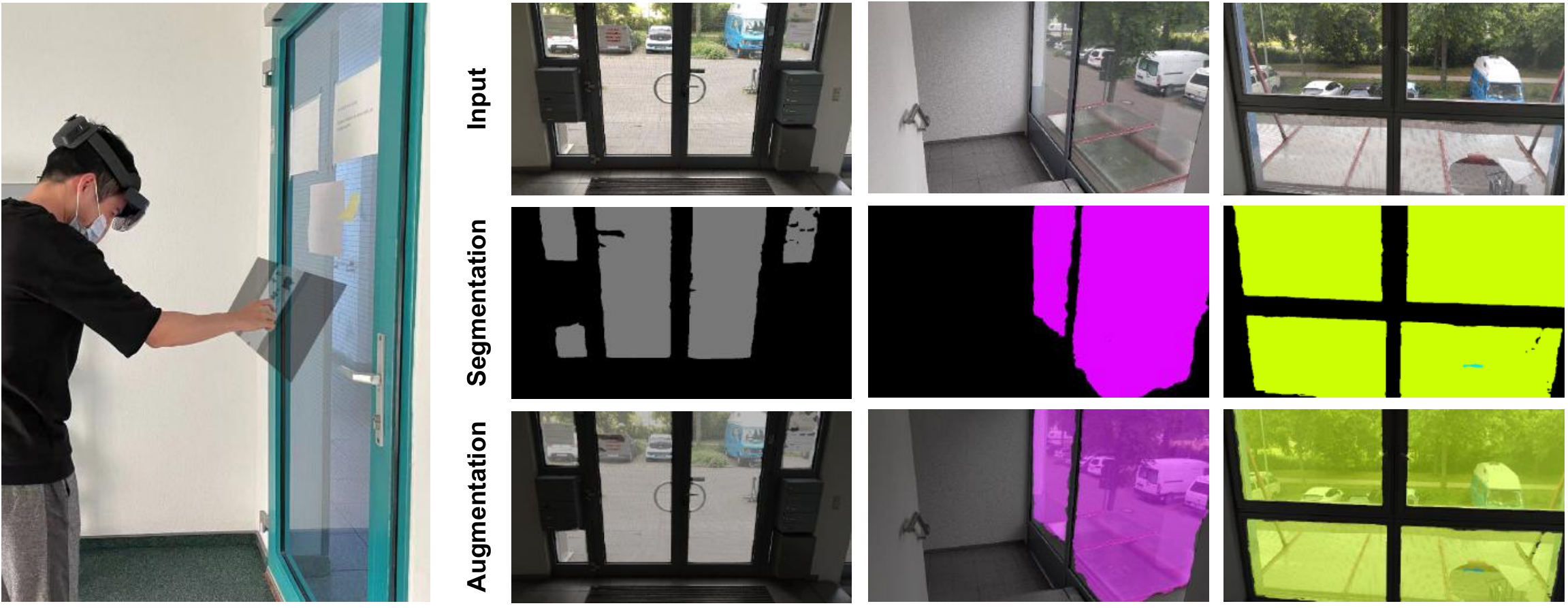}
    \caption{Augmented results with a HoloLens 2 device.}
    \label{fig:hololens}
    \vskip-2ex
\end{figure}

\noindent\textbf{Augmented reality for partially sighted people.}
As transparent obstacles are usually a threat for people with low vision and even hardly distinguishable for sighted people in some confusing situations, our Trans4Trans model are further tested on a HoloLens 2 device by capturing real-world images around our computer vision laboratory. As displayed in Fig.~\ref{fig:hololens}, the challenging and omnipresent transparent objects like \emph{glass door}, \emph{transparent wall}, and \emph{glass window} can be reliably and completely segmented, and the colored segmentation masks can be easily overlaid and naturally projected onto the original RGB images shot by the glasses for rendering augmented- or mixed reality. This field test on another glasses device reveals that our segmentation model is robust across cameras and the proposed Trasn4Trans framework can not only assist blind people, but can potentially help partially sighted people.

\section{Conclusion}
In this work, we tackle the challenges of transparent object and semantic scene segmentation via Trans4Trans, an efficient transformer architecture with both transformer-based encoder and decoder. At the heart of our assistive system is Trans4Trans, which precisely segments general- and transparent objects with a Transformer Parsing Module (TPM) integrated in the dual-head structure. It achieves state-of-the-art performances on Trans10K-v2 and Stanford2D3D datasets, while being swift and robust to support safety-critical navigation assistance. Considering the synergy between walking- and driving scene perception for improving traffic safety, Trans4Trans is further verified on driving scene segmentation benchmarks including Cityscapes (favorable conditions), ACDC (adverse conditions), and DADA-seg (extreme accident conditions), demonstrating its efficiency and robustness for real-world transportation applications.

The efficient vision transformer is ported in our wearable system with a pair of smart vision glasses designed to help visually impaired people travel and explore surrounding scenes, where transparent objects widely exist in the real life and impede their mobility. Despite a limited number of participants, an extensive set of analyses from a user study and various field tests evidences that the proposed assistive system is reliable and cognitive-load friendly.

In the future, we aim to address the issues and suggestions mentioned in the user study and improve the system continuously.
We also intend to address indoor localization for assisting visually impaired people in their navigation tasks. We plan to further look into the challenging accident scene segmentation task via the lenses of curriculum-based and open compound domain adaptation.

\bibliographystyle{IEEEtran}
\bibliography{bib}

\end{document}